\documentclass[10pt,twocolumn,letterpaper]{article}

\usepackage[pagenumbers]{wacv} 
\usepackage{threeparttable}
\usepackage{graphicx} 
\usepackage{subcaption}
\usepackage{adjustbox}
\usepackage{subcaption}
\usepackage{booktabs}
\usepackage{siunitx}
\usepackage{makecell}
\usepackage{booktabs}
\usepackage{multirow}
\usepackage{threeparttable}
\usepackage[table]{xcolor}
\usepackage{pgf}

\usepackage[
  draft,
  marginwidth=1.7cm,
  marginsep=0.15cm
]{labreview}

\definecolor{RecallDrop}{HTML}{4393C3}
\definecolor{RecallNeutral}{HTML}{F2F2F2}
\definecolor{RecallRise}{HTML}{D6604D}

\newcommand{\Rzero}{90.3575}

\newcommand{\diffcell}[2]{%
  \begingroup
  \pgfmathsetmacro{\delta}{\Rzero-(#1)}%
  \pgfmathtruncatemacro{\mix}{%
    round(min(100,100*abs(\delta)/\Rzero))%
  }%
  \ifdim\delta pt<0pt
    \edef\applycolor{%
      \noexpand\cellcolor{RecallRise!\mix!RecallNeutral}%
    }%
  \else
    \edef\applycolor{%
      \noexpand\cellcolor{RecallDrop!\mix!RecallNeutral}%
    }%
  \fi
  \applycolor
  \color{black}\strut #2%
  \endgroup
}

\newcommand{\dcell}[1]{%
  \diffcell{#1}{#1}%
}

\definecolor{CBStrongGood}{HTML}{4393C3} 
\definecolor{CBMildGood}{HTML}{92C5DE}   

\definecolor{CBNeutral}{HTML}{E6E6E6}    

\definecolor{CBMildBad}{HTML}{F4A582}    
\definecolor{CBStrongBad}{HTML}{D6604D}  

\definecolor{RecallLow}{HTML}{4393C3}      
\definecolor{RecallNeutral}{HTML}{E6E6E6}  
\definecolor{RecallHigh}{HTML}{D6604D}     

\newcommand{\recallcell}[3]{%
  \begingroup
  \pgfmathsetmacro{\cellvalue}{max(0,min(100,#1))}%
  \pgfmathsetmacro{\cellbase}{#2}%
  \ifdim\cellvalue pt<\cellbase pt
    \pgfmathtruncatemacro{\mix}{%
      max(0,min(100,round(
        100*(\cellbase-\cellvalue)/\cellbase
      )))%
    }%
    \edef\applycellcolor{%
      \noexpand\cellcolor{RecallLow!\mix!RecallNeutral}%
    }%
  \else
    \pgfmathtruncatemacro{\mix}{%
      max(0,min(100,round(
        100*(\cellvalue-\cellbase)/(100-\cellbase)
      )))%
    }%
    \edef\applycellcolor{%
      \noexpand\cellcolor{RecallHigh!\mix!RecallNeutral}%
    }%
  \fi
  \applycellcolor
  \color{black}\strut #3%
  \endgroup
}

\newcommand{\DthreeCell}[1]{%
  \recallcell{#1}{87.57}{#1}%
}
\newcommand{\DthreeBold}[1]{%
  \recallcell{#1}{87.57}{\textbf{#1}}%
}

\newcommand{\GramCell}[1]{%
  \recallcell{#1}{93.54}{#1}%
}
\newcommand{\GramBold}[1]{%
  \recallcell{#1}{93.54}{\textbf{#1}}%
}

\newcommand{\RineCell}[1]{%
  \recallcell{#1}{95.98}{#1}%
}
\newcommand{\RineBold}[1]{%
  \recallcell{#1}{95.98}{\textbf{#1}}%
}

\newcommand{\SPAICell}[1]{%
  \recallcell{#1}{84.34}{#1}%
}
\newcommand{\SPAIBold}[1]{%
  \recallcell{#1}{84.34}{\textbf{#1}}%
}

\newcommand{\PlainPPRow}[5]{%
    #1
    & \DthreeCell{#2}
    & \GramCell{#3}
    & \RineCell{#4}
    & \SPAICell{#5}%
}

\newcommand{\DthreeSourceRow}[5]{%
    #1
    & \DthreeBold{#2}
    & \GramCell{#3}
    & \RineCell{#4}
    & \SPAICell{#5}%
}

\newcommand{\GramSourceRow}[5]{%
    #1
    & \DthreeCell{#2}
    & \GramBold{#3}
    & \RineCell{#4}
    & \SPAICell{#5}%
}

\newcommand{\RineSourceRow}[5]{%
    #1
    & \DthreeCell{#2}
    & \GramCell{#3}
    & \RineBold{#4}
    & \SPAICell{#5}%
}

\newcommand{\SPAISourceRow}[5]{%
    #1
    & \DthreeCell{#2}
    & \GramCell{#3}
    & \RineCell{#4}
    & \SPAIBold{#5}%
}

\newcommand{\sourcerule}{%
    \specialrule{0.30pt}{0.08em}{0.06em}%
}

\newcommand{\familyrule}{%
    \specialrule{0.80pt}{0.18em}{0.10em}%
}

\newcommand{\todo}[1]{{\color{red}#1}}

\definecolor{wacvblue}{rgb}{0.21,0.49,0.74}
\usepackage[pagebackref,breaklinks,colorlinks,allcolors=wacvblue]{hyperref}

\def\wacvPaperID{2140} 
\def\confName{WACV}
\def\confYear{2027}

\title{Tricking Synthetic Image Detectors with
Amortized Generative Attacks}

\author{
Mohamed Awad\thanks{Corresponding author: \texttt{mohamed.awad@mbzuai.ac.ae}} \qquad
Khaled Elhady \qquad
Yasser Rohaim \qquad
Samuele Poppi \qquad
Nils Lukas\\[0.5em]
Mohamed bin Zayed University of Artificial Intelligence\\
Abu Dhabi, United Arab Emirates
}

\begin{document}
\maketitle
\begin{abstract}

Rapid advances in image synthesis have increased the need for reliable synthetic-image detectors, yet their robustness to transferable attacks remains incomplete and understudied. Existing evaluations generally examine whether an attack constructed against one detector transfers to another, but rarely isolate how its effectiveness changes when the attacked images originate from different generators. We conduct a systematic evaluation of three representative attacks: two iterative perturbation-based methods, MIG and MuMoDIG, and the amortized generative attack Tricker. Across four detectors and five generators spanning diverse architectures, we introduce a two-axis evaluation protocol that independently varies the surrogate detector used to construct or train the attack and the generator supplying its input images. We assess attack effectiveness, cross-detector transferability, perceptual quality, runtime, and interaction with common post-processing operations. We revisit Tricker and show that its data and training requirements can be substantially reduced, enabling broader and more efficient evaluation. Our results demonstrate that transferability is not a detector-only property: the input-generator distribution strongly influences both evasion effectiveness and attacked-image quality. MIG and MuMoDIG exhibit pronounced dependence on the detector--generator pair, whereas Tricker provides more uniform white-box evasion and image quality. We additionally observe that post-processing maintains the asymmetric transferability topology. These findings expose a neglected dimension of attack evaluation and motivate transfer-aware protocols that jointly account for detector and generator variation. We will release our code, evaluation data, and trained attack models with a responsible-use statement.

\end{abstract}
    
\section{Introduction}
\label{intro}

Advances in deep neural network architectures have been a key factor behind the remarkable success of recent image generation models~\cite{dhariwal2021diffusion,esser2024scaling,karras2020analyzing,rombach2022high,saharia2022photorealistic,peebles2023scalable}. These models~\cite{midjourney2025,labs2025flux,esser2024scaling} are now capable of producing high-fidelity photorealistic images that increasingly blur the boundary between real and synthetic content. As a result, developing robust detectors~\cite{corvi2023detection,karageorgiou2025any,koutlis2024leveraging,wang2020cnn,yan2024sanity,yang2025d} that can reliably distinguish authentic images from generated ones has become increasingly important for combating fake news, misinformation, and malicious use~\cite{vaccari2020deepfakes,chesney2019deep,ferrara2024genai,marchal2024generative}.

Distinguishing between unmodified synthetic and real images is insufficient by itself. In practice, detectors must remain robust under common image transformations~\cite{schinas2024sidbench,zhu2023genimage,li2025bridging,guillaro2025bias} and, more importantly, against adversarial attacks that exploit inherent vulnerabilities in neural networks~\cite{goodfellow2014explaining,saberi2023robustness,diao2024vulnerabilities}. This makes it essential to study attacks that expose weaknesses in these detectors, as doing so helps us better understand their failure modes and improve their robustness, rather than foster a false sense of security~\cite{athalye2018obfuscated,madry2017towards}, consistent with red-teaming practice~\cite{dehdashtian2025polyjuice,li2023discrete,jia2022adv}.

A wide range of threat models has been proposed in the literature, giving rise to diverse attack strategies. Black-box attacks are particularly important because they reflect realistic deployment constraints~\cite{mavali2024adversarial}. Among them, transfer-based attacks use a fully accessible surrogate model to craft samples that also deceive protected target models~\cite{diao2024vulnerabilities,neekhara2021adversarial,hou2023evading}. The key risk of this threat model lies in exploiting similarities among models, such as shared feature representations or decision boundaries, thereby demonstrating how open-source models can become a source of security risk for deployed systems.

Transferability~\cite{szegedy2013intriguing,papernot2016transferability} is generally studied with perturbation attacks~\cite{eddoubi2025raid}. However, a less explored direction is adversarial generation, where a neural transformation network is trained to map synthetic images to detector-evasive variants while preserving their visual content. This direction has already been shown to be feasible by Adversarial Reality~\cite{ciftci2025adversarial}. However, the original work provide limited analysis and less capable architecture. We therefore treat Adversarial Reality as motivation for the generative-attack paradigm and focus our comparison on reproducible transferable perturbation attacks. 


We therefore study adversarial generation through the lens of amortized transferability. In synthetic-image detection, every attack instance is defined by both the generator that produced the image and the detector being attacked. This creates a richer transfer setting than standard classifier attacks: an attack may transfer across detectors for a fixed generator, across generators for a fixed detector, or across both. We use Tricker to denote a neural transformation network trained on a source detector–generator pair. Unlike per-image perturbation attacks~\cite{madry2017towards,carlini2017towards,ma2023transferable,ren2025improving}, Tricker incurs one-time training cost and then attacks new images with a single forward pass. We investigate the following question: can an amortized Tricker trained on one source generator--detector pair produce adversarial synthetic images that transfer to unseen detector and generator, and when does this amortized strategy become preferable to state-of-the-art per-image transferable attacks? To answer that, we make the following contributions:

\begin{itemize}

\item We introduce a detector–generator evaluation protocol that reveals asymmetric transfer patterns and identifies effective source detectors, fragile targets, and isolated transfer pairs.

\item We provide a comprehensive comparison between generative attacks and strong transferable perturbation attacks, MIG and MuMoDIG, analyzing attack success, perceptual quality, transferability, per-image crafting time, break-even points, and survivability under common post-processing operations.

\item We revisit generative adversarial attacks for synthetic-image detection and show that a compact residual-attention autoencoder can be trained efficiently against a frozen surrogate detector, requiring only a small source-generator dataset while preserving image quality.

\end{itemize}
\section{Related Work}
\label{related}
Beyond additive perturbations, a less explored approach is to train a neural network to trick a synthetic-image detector. This strategy is potentially more general because it requires gradient access only to a surrogate detector during training and is not inherently tied to a particular image generator or detector architecture. This distinguishes it from methods such as PolyJuice~\cite{dehdashtian2025polyjuice}, which directly steer the image-generation process and may therefore be restricted to specific generator families or generation paradigms.

\textbf{Adversarial Reality~\cite{ciftci2025adversarial}.} This work provides an important foundation for generating adversarial samples against deepfake detectors by demonstrating the feasibility of using a conventional U-Net~\cite{ronneberger2015u} as a neural ``Tricker'' network. However, as a foundational study, it leaves substantial room for improvement. In particular, its training cost can be substantially reduced through more effective architectural choices, such as residual-attention blocks, together with a stronger training recipe. Adversarial Reality demonstrates the feasibility of a generative approach for evading deepfake detectors. Because its implementation details are insufficient for reproducibility, we treat its reported performance as evidence for the attack paradigm rather than as a directly comparable baseline. Moreover, its transferability analysis is limited, as the evaluated detectors are predominantly CNN-based and therefore do not fully capture the challenges of heterogeneous transfer or support their generalization claims. This is important because prior work has shown that transferability can be limited across different detector families, particularly between CLIP-based and CNN-based architectures~\cite{de2024exploring}.

\textbf{MIG and MuMoDIG~\cite{ma2023transferable,ren2025improving}.} They are adversarial attacks designed to improve transferability across image classifiers, including CNNs, ViTs, and even adversarially robust classifiers~\cite{awad2025defense}. MIG builds on Integrated Gradients~\cite{sundararajan2017axiomatic}, an interpretability method with relatively architecture-invariant attribution properties, to craft perturbations that better capture transferable model vulnerabilities. MuMoDIG further extends this idea by refining the integrated-gradient path along three dimensions: multiplicity, monotonicity, and diversity. These refinements are intended to produce stronger transferable perturbations and improve upon the original MIG formulation.

\textbf{RAID~\cite{eddoubi2025raid}.}
It introduces a standardized benchmark containing 72K transferable adversarial examples, constructed by attacking an ensemble of seven synthetic-image detectors on images from four generators at three $\ell_\infty$ perturbation budgets. Its precomputed examples enable fast and reproducible assessment of detector robustness against strong ensemble-based perturbations. Our study is complementary: rather than providing a fixed adversarial corpus, we introduce a two-axis evaluation protocol that varies the surrogate and target detectors and, for Tricker, the source and target generator distributions. Thus, RAID supports standardized robustness screening, whereas our protocol diagnoses how attack behavior and transferability vary across detector--generator configurations and attack mechanisms.
\section{Methodology}
\label{methodology}

The adversarial generation paradigm proposed in prior work does not fully demonstrate how to train such models in a data- and training-efficient manner. Revisiting this direction is therefore necessary to reveal the latent potential of neural adversarial generation as a practical attack strategy. In our setting, we consider a Tricker network $T$ whose goal is to transform images produced by a generator $G$ so that they deceive a target detector $D$ while preserving the content of the generated image. All experiments were conducted on an NVIDIA RTX 5000 Ada GPU. For an overview of the detectors ( GramNet~\cite{liu2020global}, RINE~\cite{koutlis2024leveraging}, $\mathrm{D}^3$~\cite{yang2025d}, SPAI~\cite{karageorgiou2025any}) and generators (BigGAN~\cite{brock2018large}, VQDM~\cite{gu2022vector}, SDV1.5~\cite{rombach2022high}, VAR~\cite{tian2024visual}, SDV3.5~\cite{esser2024scaling}), see \autoref{app:com_related_work}.

\textbf{Training.} The autoencoder is trained using a frozen SIDBench detector as a surrogate model. Each input image $x$ is resized and center-cropped to $256 \times 256$. Given an input image, the autoencoder produces a reconstructed image $\tilde{x}=T_\theta(x)$, which is then passed to the frozen surrogate detector after applying the detector-specific preprocessing. Tricker is optimized using AdamW with a learning rate of $10^{-4}$, a warm-up followed by cosine learning-rate decay, and batch size of 8. Full details are provided in \autoref{app:impl}. The training objective combines a surrogate-guided KL-divergence term with an RMSE constraint:
\[
\mathcal{L}(\theta)
=
\alpha\,D_{\mathrm{KL}}
\left(
q \,\|\, p_{D_s}(\cdot \mid \tilde{x})
\right)
+
\beta\,\mathrm{RMSE}(x,\tilde{x}),
\]
where $p_{D_s}(\cdot \mid \tilde{x})$ denotes the output distribution of the frozen surrogate detector for the reconstructed image, $q=[1,0]$ denotes the real-class target distribution, assuming class order [real, fake], and $\mathrm{RMSE}$ is root mean square error. 
This objective encourages the reconstructed image to trick the surrogate detector while remaining visually close to the original input. The best checkpoint is selected according to the validation objective, with early stopping applied when the validation loss does not improve for four consecutive epochs or when training reaches 60 epochs. Although attack success saturates early, image fidelity can continue to improve. We therefore apply the same pre-specified early-stopping rule to every Tricker run and stop optimization once the validation objective ceases to improve. 

\begin{figure}[t]
  \centering
  \begin{subfigure}[t]{.48\columnwidth}
    \centering
    \includegraphics[width=\linewidth]{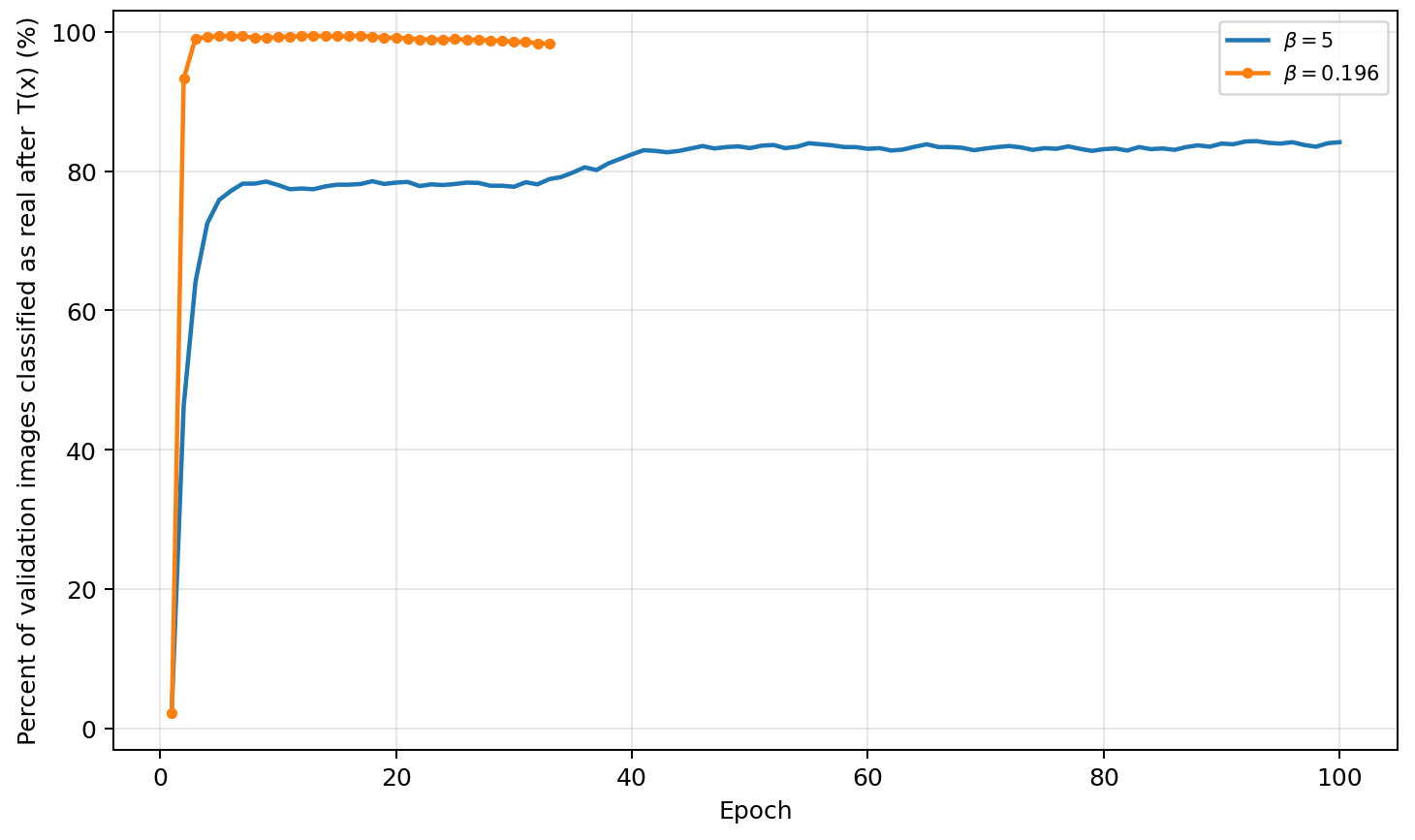}
    \caption{Attack success rate on validation images.}
    \label{fig:ASR_diff_beta}
  \end{subfigure}
  \begin{subfigure}[t]{.48\columnwidth}
    \centering
    \includegraphics[width=\linewidth]{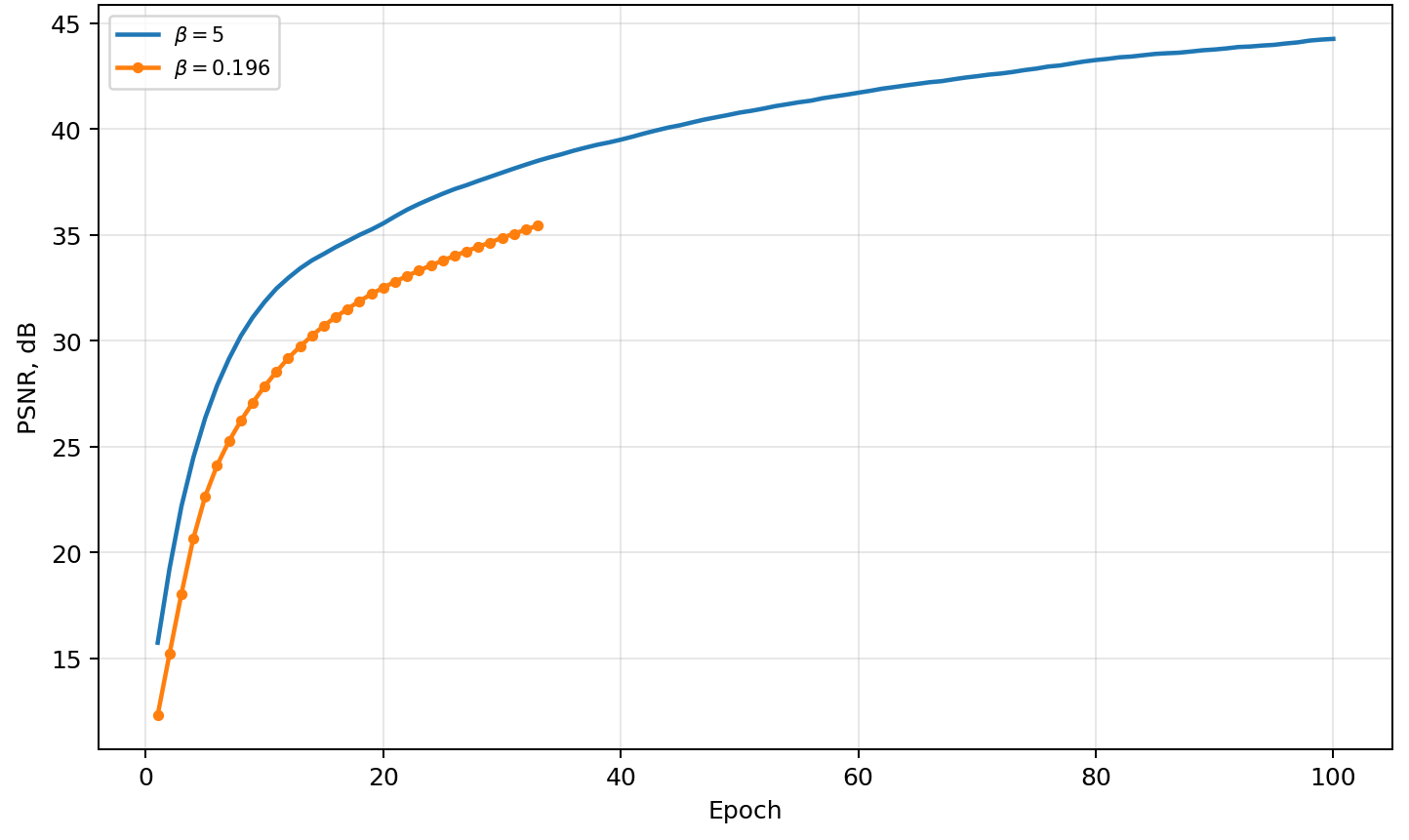}
    \caption{PSNR for reconstructed validation images.}
    \label{fig:PSNR_diff_beta}
  \end{subfigure}
  \caption{The effect of scaling the reconstruction term in the loss.}
  \label{fig:compare_betas}
\end{figure}

Fig.~\ref{fig:compare_betas} shows the effect of varying $\beta$ while keeping $\alpha$ fixed for Tricker trained with RINE–BigGAN. With smaller values of $\beta$, the attack success rate rapidly reaches nearly 99\% by the third epoch. In contrast, larger values of $\beta$ cause the attack success rate to plateau around 82\%, which is close to the average white-box attack success rate reported in Adversarial Reality~\cite{ciftci2025adversarial}. Importantly, achieving a high attack success rate does not come at a substantial cost to image quality, as shown in Fig.~\ref{fig:compare_betas}. The PSNR continues to improve during training, suggesting that further quality gains may be possible; however, in our setup, training is terminated by early stopping once the attack objective has converged.

\textbf{Tricker.} We build a residual-attention U-Net-style autoencoder~\cite{ronneberger2015u}, consisting of an encoder–decoder structure with skip connections. The architecture is designed to preserve fine spatial details through encoder–decoder feature concatenation while also capturing long-range visual dependencies using attention blocks. The encoder contains three residual stages with channel widths of 32, 64, and 128, where each stage consists of two residual blocks using Group Normalization, SiLU activation, and 3×3 convolutional layers, followed by stride-2 convolutional downsampling. The deepest representation is formed using a 256-channel residual stage followed by a bottleneck module composed of residual blocks and a self-attention block. Attention is applied at the higher-capacity levels, including the 128-channel encoder stage, the 256-channel deep stage, the bottleneck, and the first decoder stage. The decoder mirrors the encoder by bilinearly upsampling the feature maps, concatenating them with the corresponding encoder features through skip connections, and refining them using residual stages. A final residual head, followed by Group Normalization, SiLU activation, a 3×3 convolution, and sigmoid activation, produces the reconstructed RGB image in the range [0, 1]. With the default configuration of 32 base channels, channel multipliers of 1, 2, 4, and 8, and two residual blocks per stage, the architecture contains about 8 million parameters.

\begin{figure}[t]
  \centering
  \begin{subfigure}[t]{.48\columnwidth}
    \centering
    \includegraphics[width=\linewidth]{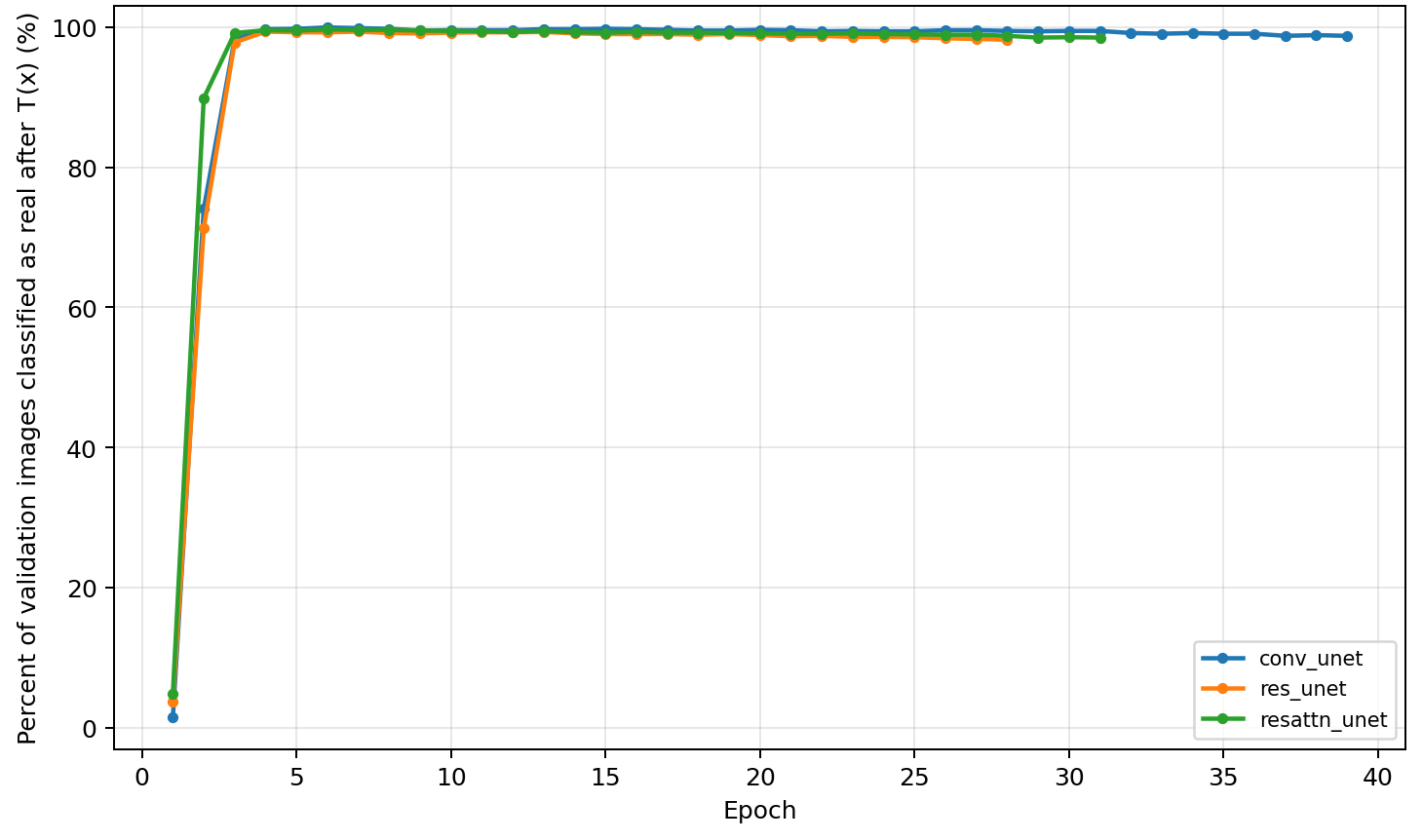}
    \caption{Attack success rate on validation images.}
    \label{fig:ASR_diff_arch}
  \end{subfigure}
  \begin{subfigure}[t]{.48\columnwidth}
    \centering
    \includegraphics[width=\linewidth]{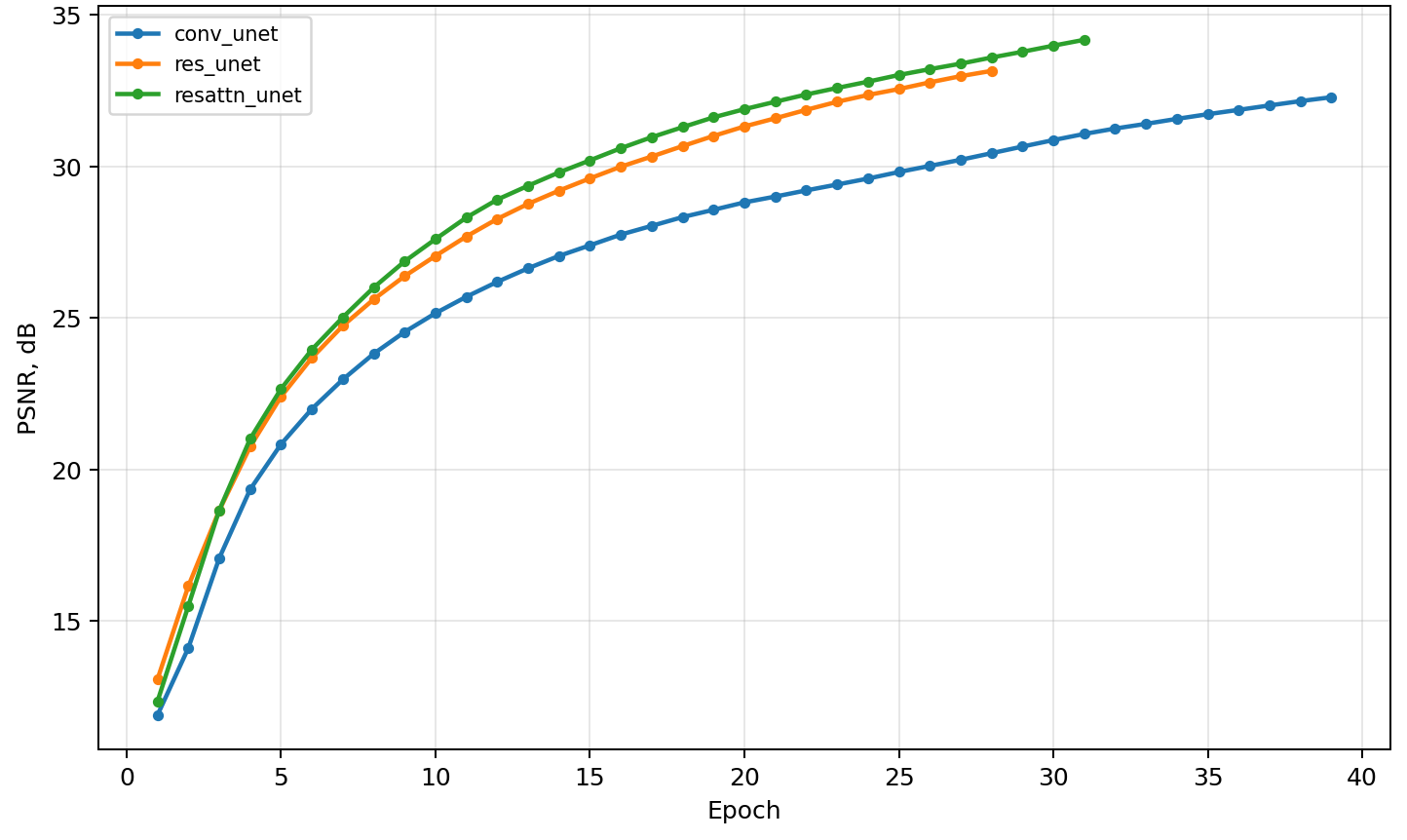}
    \caption{PSNR for reconstructed validation images.}
    \label{fig:PSNR_diff_arch}
  \end{subfigure}
  \caption{Ablation study for selecting the U-Net architecture.}
  \label{fig:compare_arch}
\end{figure}

To justify our architectural choice, we ablate three U-Net variants with approximately matched capacity: a convolutional U-Net baseline, following the architecture used in prior work; U-Net with residual connections; and our residual-attention U-Net, as shown in Fig.~\ref{fig:compare_arch}. All variants are trained with RINE–BigGAN using the same recipe and a low value of $\beta$. Under this setting, all models reach nearly $100\%$ ASR by the third epoch, suggesting that high attack success is primarily driven by the choice of $\beta$. However, the architectures differ substantially in image fidelity. Our residual-attention U-Net achieves the highest PSNR at convergence, indicating better preservation of image quality. The residual U-Net performs competitively, with only a small gap to the attention-based variant, whereas the purely convolutional baseline lags behind despite requiring more epochs to converge. 

\textbf{Data.} We primarily use GenImage~\cite{zhu2023genimage}, either by directly using its released synthetic images or by following its protocol to generate new images from recent generative models. Although GenImage covers eight generators, some of them share similar architectural designs and generation mechanisms, such as SDV1.4 and SDV1.5, and therefore do not fully capture the diversity of modern image generation paradigms, including visual autoregressive models and flow-matching-based generators~\cite{lipman2022flow}. To obtain a more diverse evaluation set, we directly use BigGAN, SDV1.5, and VQDM images from GenImage, and additionally generate images from VAR and SDV3.5 following the GenImage data construction protocol. In total, we prepare 30,000 synthetic images from five generators, with 6,000 images per generator. For each generator, 4,000 images are used for training and 2,000 for attacker evaluation.

Similarly, we seek to cover diverse detection paradigms. We therefore select four detectors and build on the SIDBench codebase~\cite{schinas2024sidbench}, which provides a unified framework for evaluating a wide range of synthetic image detectors. Among the detectors already supported by SIDBench, RINE and GramNet achieve notably high recall, so we include them as representative baselines. To expand the benchmark with more recent state-of-the-art methods, we additionally integrate two detectors, $\mathrm{D}^3$ and SPAI, into the same framework. For all detectors, we carefully preserve their required preprocessing pipelines, including resizing, cropping, and normalization procedures, while maintaining a consistent evaluation setup. This integration ensures that performance differences reflect detector behavior rather than implementation artifacts or preprocessing mismatches.

\textbf{Threat Model.} We evaluate attacks under two access settings: white-box and transfer-based black-box. In the white-box setting, the attacker has gradient access to the evaluated detector. In the transfer-based black-box setting, the target detector is never queried; the attacker has gradient access only to a surrogate detector $D_s$.
For Tricker, the source generator $G_s$ denotes the image distribution used to train the transformation network and does not imply access to the internals of the target generator. For perturbation attacks, which are optimized independently for each input image, generator labels denote the input distribution rather than a trained generator-level surrogate. Our primary objective is to assess whether an attack can flip synthetic images into the real class with respect to $D_t$. Therefore, we report recall at a fixed threshold of 0.5 to quantify the impact of the attack on fake-image detection performance. For MIG and MuMoDIG, we use $\epsilon = 16/255$ in $l_\infty$ norm, following the strong-transfer settings; all other hyperparameters are set to default values in TorchAttack v1.7.2~\cite{woo2023torchattack}.

For clarity, we denote by $D_s$ and $D_t$ the source and target detectors, and by $G_s$ and $G_t$ the source and target generators. In the white-box setting, $D_s = D_t$ and $G_s = G_t$. In detector transfer, $D_s \neq D_t$, while the generator may be matched or unmatched. With the detector fixed, generator transfer applies only to Tricker: it is trained on $G_s$ and evaluated on $G_t \neq G_s$. MIG and MuMoDIG are optimized independently for each target image; thus, the generator label indicates only the target-image distribution.


\begin{figure}[b]
  \centering
  \begin{subfigure}[t]{.48\columnwidth}
    \centering
    \includegraphics[width=\linewidth]{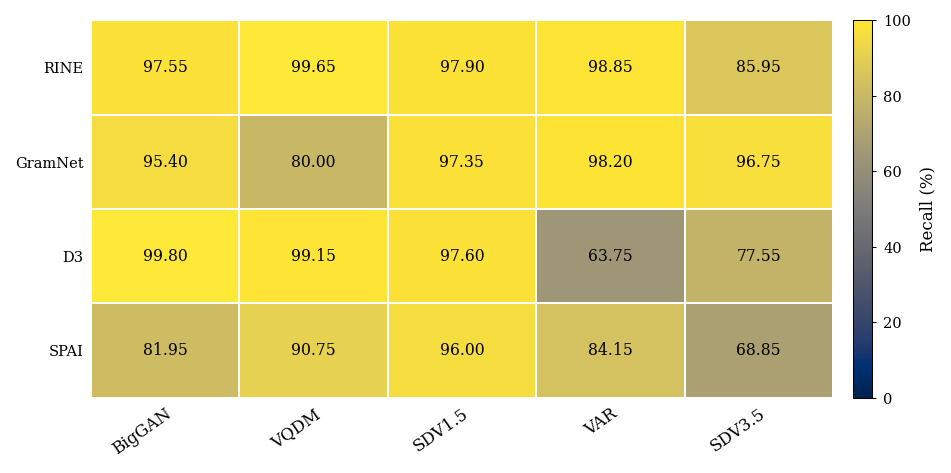}
    \caption{Original recall}
    \label{fig:before_attack}
  \end{subfigure}
  \begin{subfigure}[t]{.48\columnwidth}
    \centering
    \includegraphics[width=\linewidth]{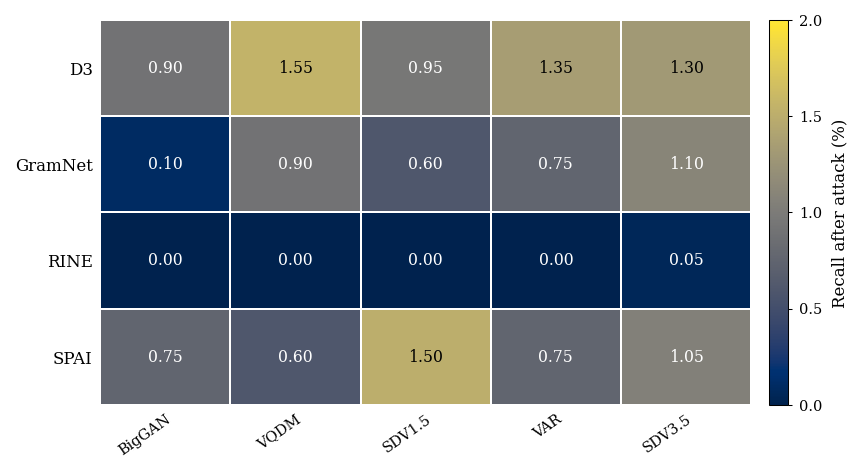}
    \caption{Tricker}
    \label{fig:white_box_tricker}
  \end{subfigure}
  \begin{subfigure}[t]{.48\columnwidth}
    \centering
    \includegraphics[width=\linewidth]{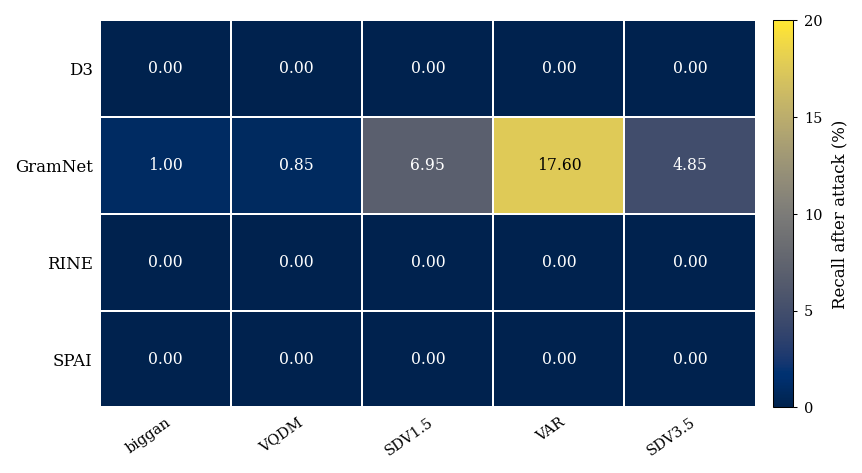}
    \caption{MIG}
    \label{fig:white_box_mig}
  \end{subfigure}
  \begin{subfigure}[t]{.48\columnwidth}
    \centering
    \includegraphics[width=\linewidth]{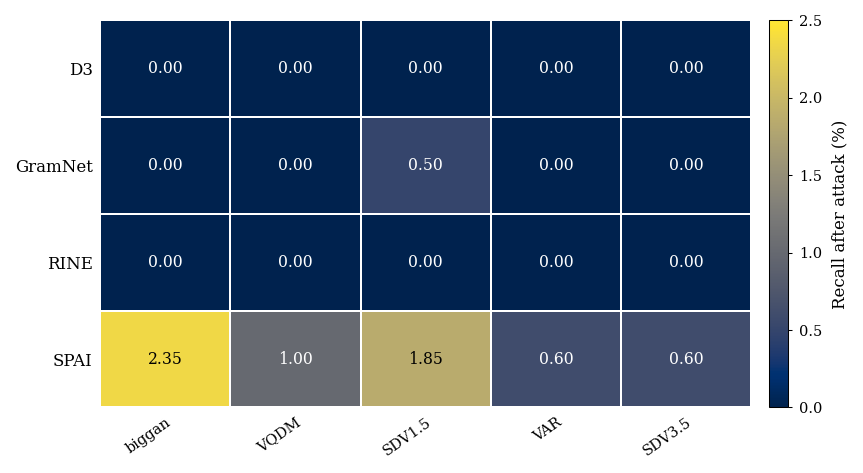}
    \caption{MuMoDIG}
    \label{fig:white_box_mumodig}
  \end{subfigure}
  \caption{
  Fake-image recall before attack (a) and under white-box attacks (b, c, d). Lower recall indicates stronger attack success. Color scales differ across panels to reveal low-recall differences.}
  \label{fig:white_box_setting}
\end{figure}
\section{Results}
\label{results}
We evaluate generative and perturbation attacks along four axes: detector evasion, perceptual fidelity, attack-generation time and survivability after post-processing. Detector evasion is measured by the drop in fake-image recall under white-box and transfer settings, while perceptual fidelity is measured using PSNR, SSIM, and RMSE. Recall denotes fake-class recall at a fixed threshold of 0.5. Since the attack objective is to make synthetic images classified as real, lower recall corresponds to stronger attack success. Clean detector recall before attack is shown in Fig.~\ref{fig:before_attack}; results for the full SIDBench detector set are provided in \autoref{app:base}. 

\sisetup{
  table-number-alignment = center,
  detect-weight = true,
  detect-family = true
}

\begin{table*}[t]
\centering
\caption{Image-quality comparison in terms of PSNR, RMSE, SSIM, LPIPS, and FID across white-box attacks. Results are averaged over the five image generators. Each detector-level value is averaged over generators. Higher PSNR and SSIM indicate better image quality, whereas lower RMSE, LPIPS, and FID are better. Boldface denotes the best detector-specific result within each attack and metric. The final row reports the mean $\pm$ sample standard deviation over the four detector settings.}
\label{tab:quality_metrics_white_box}

\begin{threeparttable}
\footnotesize
\setlength{\tabcolsep}{1.35pt}
\renewcommand{\arraystretch}{1.10}

\sisetup{detect-weight=true}

\newcommand{\avgstd}[2]{%
    \multicolumn{1}{c}{%
        \makecell[c]{#1\\[-1pt]\scriptsize$\pm$\,#2}%
    }%
}

\begin{adjustbox}{max width=\textwidth}
\begin{tabular}{@{}l
S[table-format=2.2,detect-weight=true]
S[table-format=2.2,detect-weight=true]
S[table-format=1.4,detect-weight=true]
S[table-format=1.4,detect-weight=true]
S[table-format=2.2,detect-weight=true]
S[table-format=2.2,detect-weight=true]
S[table-format=2.2,detect-weight=true]
S[table-format=1.4,detect-weight=true]
S[table-format=1.4,detect-weight=true]
S[table-format=2.2,detect-weight=true]
S[table-format=2.2,detect-weight=true]
S[table-format=2.2,detect-weight=true]
S[table-format=1.4,detect-weight=true]
S[table-format=1.4,detect-weight=true]
S[table-format=2.2,detect-weight=true]
@{}}
\toprule

& \multicolumn{5}{c}{\textbf{Tricker}}
& \multicolumn{5}{c}{\textbf{MIG}}
& \multicolumn{5}{c}{\textbf{MuMoDIG}} \\

\cmidrule(lr){2-6}
\cmidrule(lr){7-11}
\cmidrule(l){12-16}

\textbf{Detector}
& {\textbf{PSNR}}
& {\textbf{RMSE}}
& {\textbf{SSIM}}
& {\textbf{LPIPS}}
& {\textbf{FID}}
& {\textbf{PSNR}}
& {\textbf{RMSE}}
& {\textbf{SSIM}}
& {\textbf{LPIPS}}
& {\textbf{FID}}
& {\textbf{PSNR}}
& {\textbf{RMSE}}
& {\textbf{SSIM}}
& {\textbf{LPIPS}}
& {\textbf{FID}} \\

\midrule

$\mathrm{D}^{3}$
& 32.41 & 6.80 & 0.9373 & 0.0621 & 12.15
& 28.22 & 9.91 & 0.7519 & 0.2902 & 26.08
& 28.30 & 9.82 & 0.7486 & 0.2973 & 24.19 \\

GramNet
& 33.16 & 6.38 & 0.9513 & 0.0352 & 7.38
& \bfseries 45.48
& \bfseries 1.89
& \bfseries 0.9800
& \bfseries 0.0140
& \bfseries 5.45
& \bfseries 45.12
& \bfseries 1.54
& \bfseries 0.9857
& \bfseries 0.0088
& \bfseries 4.21 \\

RINE
& \bfseries 37.94
& \bfseries 3.61
& \bfseries 0.9683
& \bfseries 0.0229
& \bfseries 5.96
& 27.96 & 10.22 & 0.7491 & 0.3115 & 29.02
& 28.20 & 9.89 & 0.7494 & 0.2921 & 27.48 \\

SPAI
& 33.59 & 6.16 & 0.9578 & 0.0431 & 8.76
& 31.53 & 6.91 & 0.8358 & 0.1354 & 17.83
& 43.83 & 1.71 & 0.9837 & 0.0165 & 4.54 \\

\midrule

{\makecell[l]{Avg.\\[-1pt]\scriptsize$\pm$ Std.}}
& \avgstd{34.28}{2.49}
& \avgstd{5.74}{1.44}
& \avgstd{0.9537}{0.0130}
& \avgstd{0.0408}{0.0164}
& \avgstd{8.56}{2.65}

& \avgstd{33.30}{8.28}
& \avgstd{7.23}{3.86}
& \avgstd{0.8292}{0.1083}
& \avgstd{0.1878}{0.1399}
& \avgstd{19.59}{10.55}

& \avgstd{36.36}{9.38}
& \avgstd{5.74}{4.75}
& \avgstd{0.8669}{0.1361}
& \avgstd{0.1537}{0.1629}
& \avgstd{15.11}{12.46} \\

\bottomrule
\end{tabular}
\end{adjustbox}


\end{threeparttable}
\end{table*}

\subsection{White-box setting}

We observe a severe collapse in recall under white-box setting, as shown in ~\cref{fig:white_box_tricker,fig:white_box_mig,fig:white_box_mumodig}. Across 20 detector--generator pairs, Tricker achieves an average recall of only $0.71\%$, which is comparable to the strongest perturbation-based attacks, MIG and MuMoDIG, with average recalls of $1.56\%$ and $0.35\%$, respectively. This shows that Tricker is not merely effective in isolated cases, but reaches the same near-complete detector evasion regime as high budget perturbation attacks with stable performance across pairs. Nevertheless, the results also reveal a clear pairwise dependency: some detectors, such as RINE, are fully deceived across all generators, while specific combinations retain partial robustness, e.g., GramNet against VAR under MIG has 17.6\% recall. These findings suggest that white-box robustness in synthetic-image detection is governed by the interaction between detector architecture and generator artifacts, rather than by either component alone. To position with the baseline white-box attack PGD~\cite{madry2017towards}, we report the average recall with 2.9\% for details see \autoref{app:pgd}. 

Attack success alone is not sufficient for evaluating adversarial synthetic images. Indeed, simple image transformations can often induce detector failures, but they may also introduce visually obvious manipulations. Therefore, we additionally assess the perceptual quality of the adversarial images. As shown in Table~\ref{tab:quality_metrics_white_box}, image quality varies substantially with the target detector. 
On average, Tricker achieves the highest SSIM of $0.9537$ and ties for the lowest RMSE with $5.74$, while ranking second in PSNR with $34.28$ dB. For high level perceptual metrics, Tricker achieves the best average for LPIPS and FID by a noticeable margin over MIG and MuMoDIG. More importantly, Tricker shows substantially lower variance across detectors, indicating more stable perceptual quality than perturbation-based attacks. For visual inspection, we provide samples with detectors that yield the best image quality under each attack in Fig.~\ref{fig:qualitative-best-cases}, with additional artifact analysis included in \autoref{app:artifacts}. 


\begin{table*}[t]
\centering
\caption{Runtime comparison across detectors and attack mechanisms.}
\label{tab:runtime_comparison}
\footnotesize
\setlength{\tabcolsep}{5pt}
\renewcommand{\arraystretch}{1.12}
\begin{tabular}{@{}lccccccc@{}}
\toprule
 &
\multicolumn{3}{c}{\textbf{Tricker}} &
\multicolumn{2}{c}{\textbf{MIG}} &
\multicolumn{2}{c}{\textbf{MuMoDIG}} \\
\cmidrule(lr){2-4}
\cmidrule(lr){5-6}
\cmidrule(lr){7-8}
\textbf{Detector}&
\shortstack{\textbf{Training}\\\textbf{time}\\\textbf{(s/epoch)}} &
\shortstack{\textbf{Avg.}\\\textbf{epochs}} &
\shortstack{\textbf{Attack gen.}\\\textbf{time (ms)}} &
\shortstack{\textbf{Gen.}\\\textbf{time (s)}} &
\shortstack{\textbf{Break-even}\\\textbf{(\# images)}} &
\shortstack{\textbf{Gen.}\\\textbf{time (s)}} &
\shortstack{\textbf{Break-even}\\\textbf{(\# images)}} \\
\midrule
$\mathrm{D}^3$      & 292          & 32.8 &  13.20 & 15.78         &  608 & 5.55          &  1730 \\
GramNet & 171 & 16.6  & 13.21          & 0.37 &  7956 & 0.49 & 5954 \\
RINE    & 230          & 43.8  & 13.19 & 7.79          & 1296 & 2.16    & 4693 \\
SPAI    & 602          & 19.4 & 13.19 & 8.85          & 1322 & 2.35          &  4998 \\
\bottomrule
\end{tabular}
\end{table*}

\begin{figure}[b]
    \centering
    \begingroup
    \setlength{\tabcolsep}{0pt}
    \renewcommand{\arraystretch}{0}
    \scriptsize
    \newcommand{\qualimgw}{0.195\linewidth}
    \newcommand{\qimg}[1]{\includegraphics[width=\qualimgw]{figs/#1}}
    \newcommand{\rowlab}[1]{%
        \raisebox{\dimexpr\qualimgw/2-\height/2\relax}{\rotatebox{90}{\textbf{#1}}}%
    }

    \begin{tabular}{@{}r@{\hspace{3.pt}}ccccc@{}}
        & \textbf{BigGAN} & \textbf{SDV1.5} & \textbf{SDV3.5} & \textbf{VAR} & \textbf{VQDM} \\

        \rowlab{Original} &
        \qimg{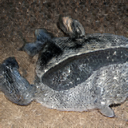} &
        \qimg{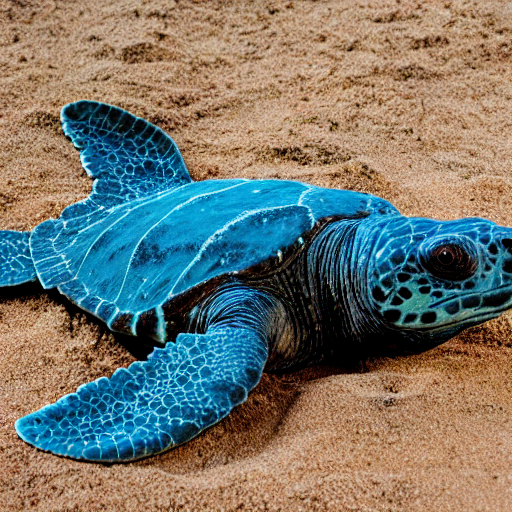} &
        \qimg{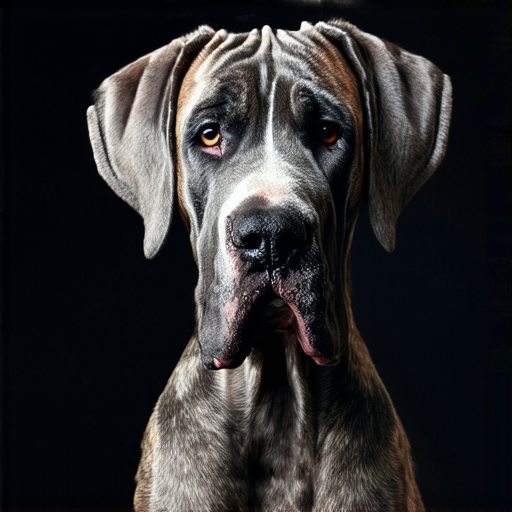} &
        \qimg{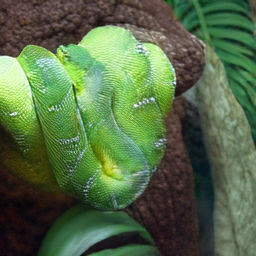} &
        \qimg{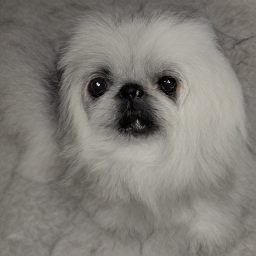} \\

        \rowlab{Tricker} &
        \qimg{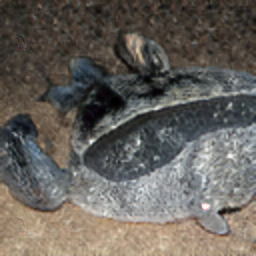} &
        \qimg{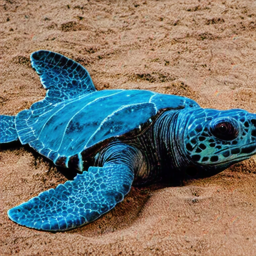} &
        \qimg{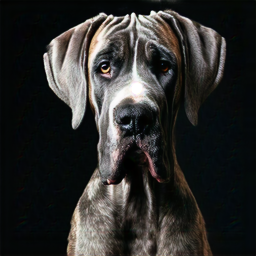} &
        \qimg{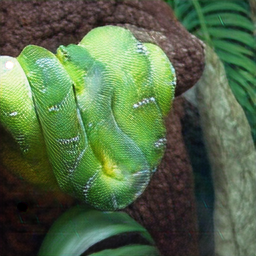} &
        \qimg{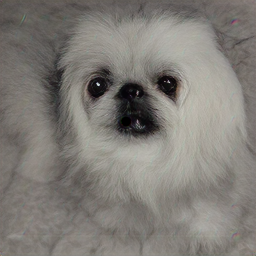} \\

        \rowlab{MIG} &
        \qimg{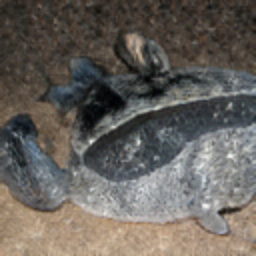} &
        \qimg{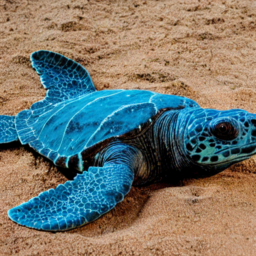} &
        \qimg{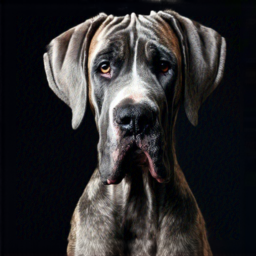} &
        \qimg{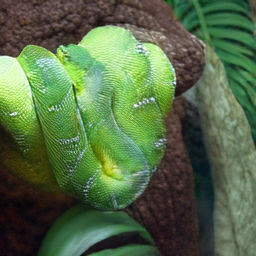} &
        \qimg{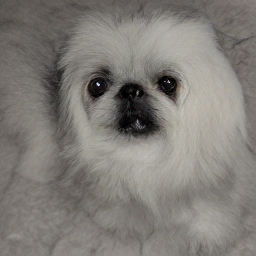} \\

        \rowlab{MuMoDIG} &
        \qimg{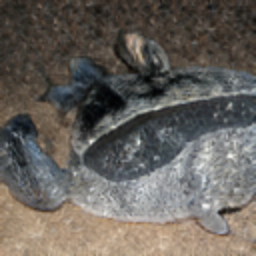} &
        \qimg{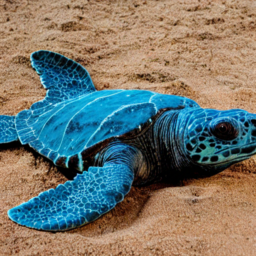} &
        \qimg{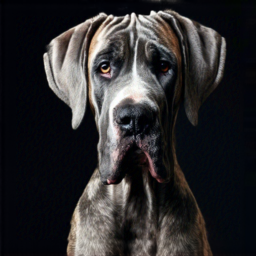} &
        \qimg{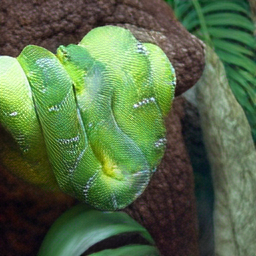} &
        \qimg{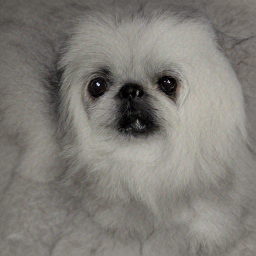}
    \end{tabular}
    \endgroup

    \caption{Qualitative comparison of original and adversarial synthetic images. Tricker examples are generated using a model trained against RINE, while MIG and MuMoDIG perturbations are crafted using GramNet as the surrogate.}
    \label{fig:qualitative-best-cases}
\end{figure}

\subsection{Amortized runtime analysis}
The time required to craft an adversarial image depends strongly on the target detector architecture and, in particular, on the cost of computing gradients through the detector. We therefore compare the runtime of each attack separately for every detector, while averaging over generators. This evaluation separates the one-time training cost of Tricker from its per-image generation cost.

Once trained, Tricker is substantially faster than iterative perturbation-based attacks. Compared with MIG, Tricker achieves speedups ranging from $28.00\times$ on GramNet to $1195.45\times$ on $\mathrm{D}^3$. Similarly, compared with MuMoDIG, Tricker is faster by $37.09\times$ on GramNet and up to $420.45\times$ on $\mathrm{D}^3$. These results show that Tricker is especially advantageous for computationally expensive detectors, where gradient-based optimization becomes the dominant cost.

The main overhead of Tricker is its training stage. With our training strategy, this one-time cost remains below $2.5$ hours on average. To quantify when this cost is amortized, we report the break-even point: the minimum number of adversarial images after which using Tricker becomes more time-efficient than an iterative attack. Formally, for a baseline attack $A$, the break-even point can be written as
\[
N_{\mathrm{BE}} = \left\lceil
\frac{T_{\mathrm{train}}^{\mathrm{Tricker}}}
{t_A - t_{\mathrm{Tricker}}} \right\rceil
\]
where $T_{\mathrm{train}}^{\mathrm{Tricker}}$ is the one-time training cost, and $t_A$ and $t_{\mathrm{Tricker}}$ are the per-image generation times of the baseline and Tricker, respectively.

This analysis reveals a clear trade-off. For lightweight detectors such as GramNet, where gradients are cheap to compute, MuMoDIG can be preferable for small attack budgets, especially since it already achieves near-zero recall. In contrast, for more complex detectors such as $\mathrm{D}^3$, the high cost of iterative gradient-based optimization leads to much smaller break-even points, making Tricker the more efficient choice when generating adversarial images at scale.


\subsection{Transferability Across Generators}

We analyze the effect of the surrogate generator on Tricker, considering both recall reduction and adversarial image quality. This experiment is only meaningful for Tricker, since it learns a reusable attack model from a surrogate generator. In contrast, perturbation-based attacks are optimized separately for each input image and do not involve a generator-level surrogate. For these methods, we instead analyze transfer and robustness across detectors.

\begin{figure}
    \centering
    \includegraphics[width=\linewidth]{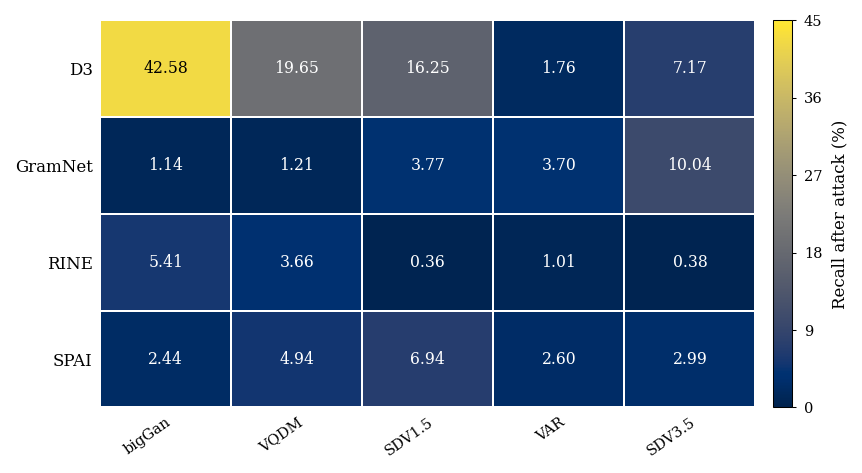}
    \caption{Average recall when Tricker is trained on all non-target source generators and evaluated on the held-out target generator for each detector.}
    \label{fig:recall_across_generator}
\end{figure}

Fig.~\ref{fig:recall_across_generator} evaluates whether an attack trained using one source generator can transfer to another target generator. Each cell reports the average recall obtained when attacking $G_t$ using Tricker models trained on the same detector but on all other source generators $G_s \neq G_t$. In other words, this is a leave-one-generator-out evaluation: the attack is never trained using images from the target generator.

For example, the $\mathrm{D}^3$--BigGAN cell reports the average recall on BigGAN images when using Tricker models trained with detector $\mathrm{D}^3$ and the source generators \{VQDM, SDV1.5, VAR, SDV3.5\}. This yields a recall of $42.58\%$, indicating limited transferability to BigGAN under $\mathrm{D}^3$. Lower recall values correspond to stronger cross-generator transfer, meaning that the target generator can be attacked effectively without access to its generated images. Conversely, higher recall values indicate weaker transferability and greater robustness for that detector--generator pair, as observed for $\mathrm{D}^3$--BigGAN. Overall, we find that in most cases Tricker transfers successfully across generators: even without training on images from the target generator $G_t$, attacks trained on alternative generators $G_s$ can substantially reduce recall on $G_t$. This suggests that Tricker often learns detector-specific vulnerabilities that are shared across generator distributions, rather than overfitting to artifacts of a single generator.

\begin{figure*}[t]
    \centering

    \begin{subfigure}[t]{0.32\textwidth}
        \centering
        \includegraphics[width=\linewidth]
        {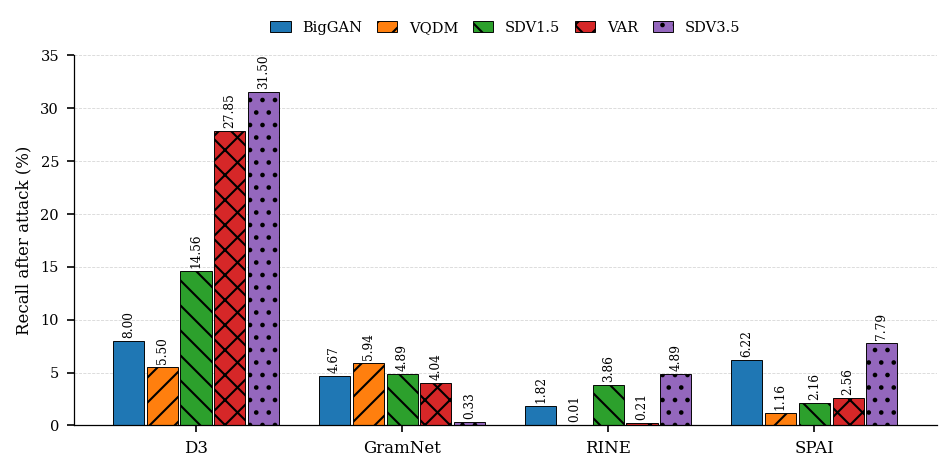}
        \caption{}
        \label{fig:surrogate_generator_recall}
    \end{subfigure}
    \hfill
    \begin{subfigure}[t]{0.32\textwidth}
        \centering
        \includegraphics[width=\linewidth]
        {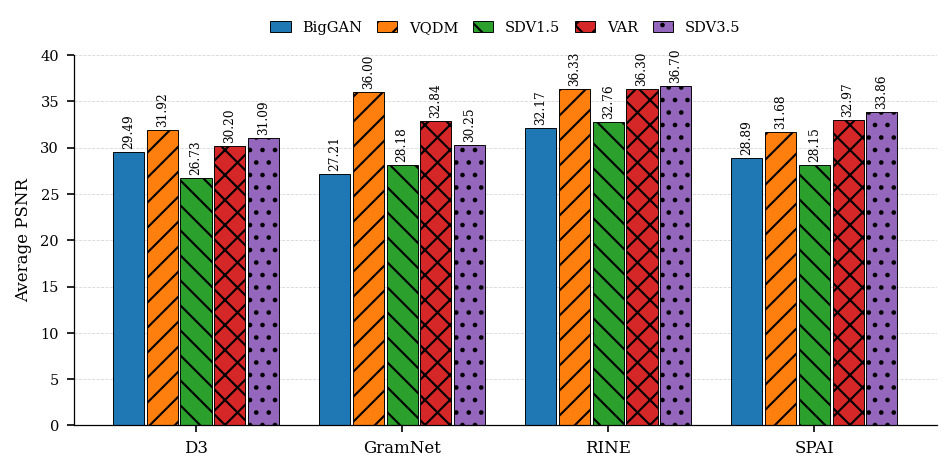}
        \caption{}
        \label{fig:surrogate_generator_PSNR}
    \end{subfigure}
    \hfill
    \begin{subfigure}[t]{0.32\textwidth}
        \centering
        \includegraphics[width=\linewidth]
        {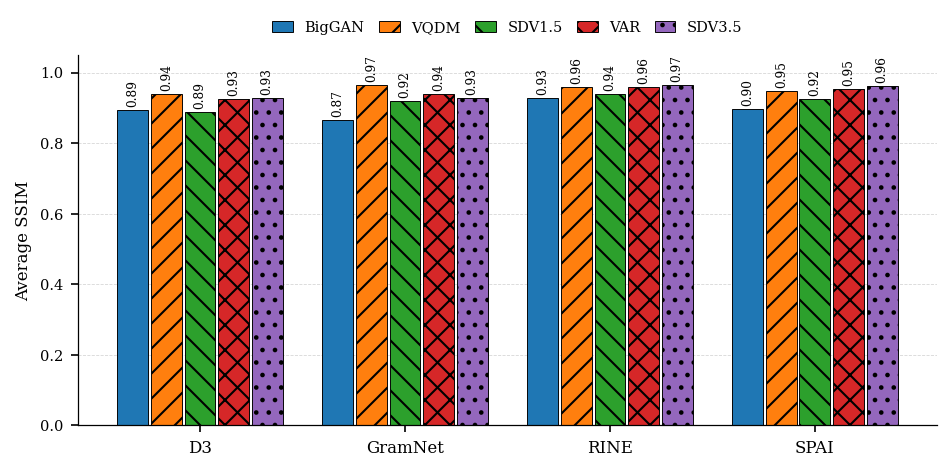}
        \caption{}
        \label{fig:surrogate_generator_SSIM}
    \end{subfigure}

    \par\vspace{1.5mm}

    \begin{subfigure}[t]{0.32\textwidth}
        \centering
        \includegraphics[width=\linewidth]
        {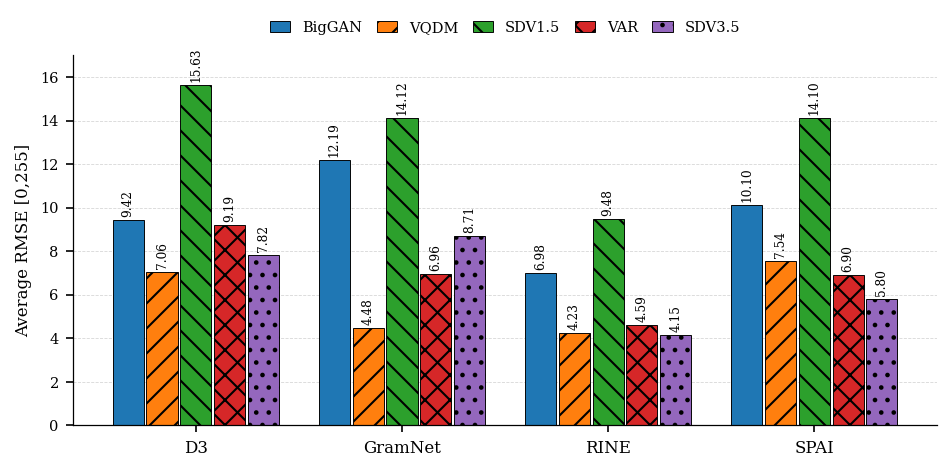}
        \caption{}
        \label{fig:surrogate_generator_RMSE}
    \end{subfigure}
    \hfill
    \begin{subfigure}[t]{0.32\textwidth}
        \centering
        \includegraphics[width=\linewidth]
        {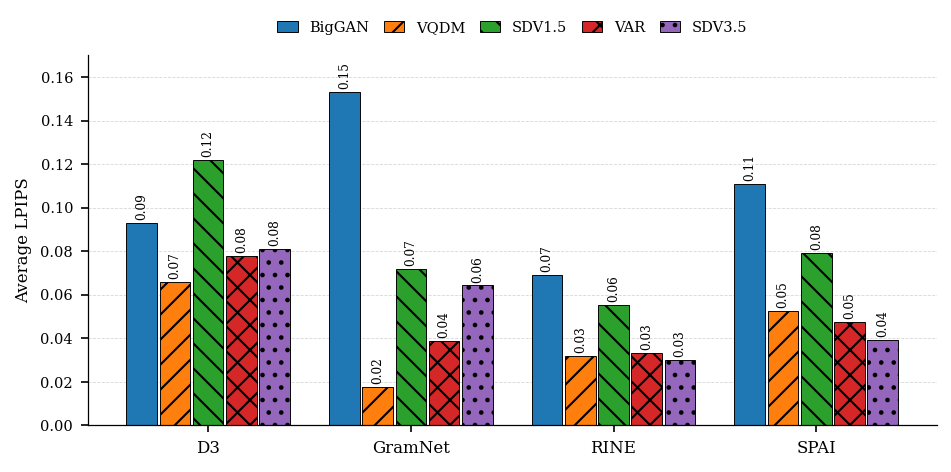}
        \caption{}
        \label{fig:surrogate_generator_LPIPS}
    \end{subfigure}
    \hfill
    \begin{subfigure}[t]{0.32\textwidth}
        \centering
        \includegraphics[width=\linewidth]
        {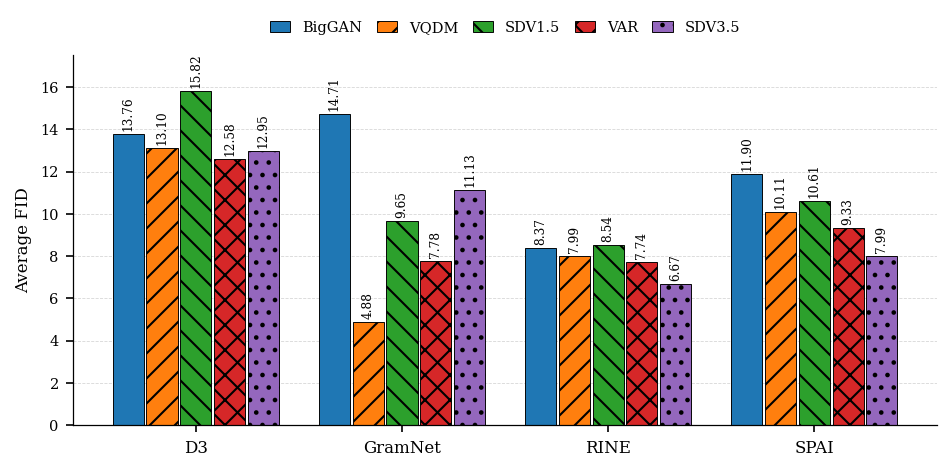}
        \caption{}
        \label{fig:surrogate_generator_FID}
    \end{subfigure}

    \caption{
        Effect of the source generator on Tricker transferability.
        For each source generator, we evaluate Tricker on unseen target
        generators using the same detector. VQDM yields the strongest
        average transfer~(a), while maintaining competitive image quality
        in terms of PSNR~(b), SSIM~(c), RMSE~(d), LPIPS~(e), and FID~(f).
    }
    \label{fig:surrogate_generator}
\end{figure*}

Fig.~\ref{fig:surrogate_generator_recall} reports average recall for each detector $D$ when Tricker is trained on the same detector with a surrogate generator $G_s$ and evaluated on the remaining generators $G_t \neq G_s$. Overall, VQDM is the strongest surrogate, leading to the largest recall reductions on average across detectors. Interestingly, ~\cref{fig:surrogate_generator_PSNR,fig:surrogate_generator_SSIM,fig:surrogate_generator_RMSE,fig:surrogate_generator_LPIPS,fig:surrogate_generator_FID} show that the surrogate generator producing the strongest recall drop also tends to yield better perceptual quality. Thus, improved transferability does not necessarily require sacrificing image fidelity. In fact, adversarial images generated for unseen target generators can sometimes achieve higher quality than those in white-box setting. For instance, GramNet obtains average PSNR of $33.16$ dB in white-box setting, while using VQDM as surrogate increases the average PSNR to $36.00$ dB.


\subsection{Transferability Across Detectors}
In all transfer heatmaps, rows denote the surrogate detector used to craft or train the attack, columns denote the held-out target detector, and lower recall means stronger transfer.

\begin{figure}[t]
  \centering
  \begin{subfigure}[t]{.48\columnwidth}
    \centering
    \includegraphics[width=\linewidth]{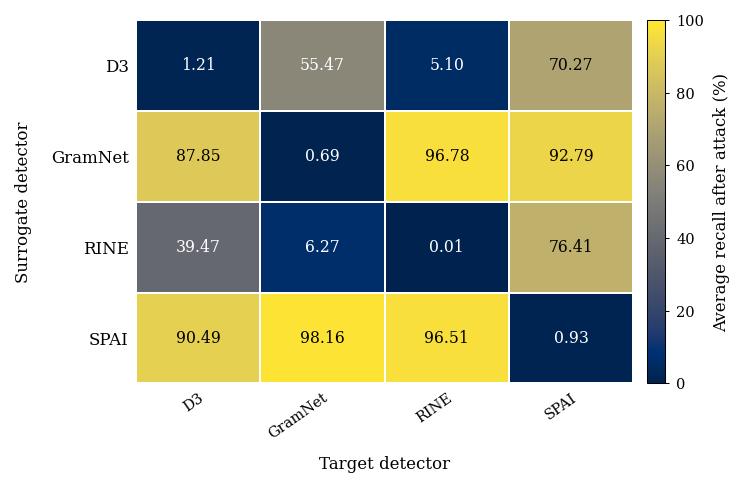}
    \caption{Tricker, matched generators}
    \label{fig:transfer_detectors_tricker_match}
  \end{subfigure}
  \begin{subfigure}[t]{.48\columnwidth}
    \centering
    \includegraphics[width=\linewidth]{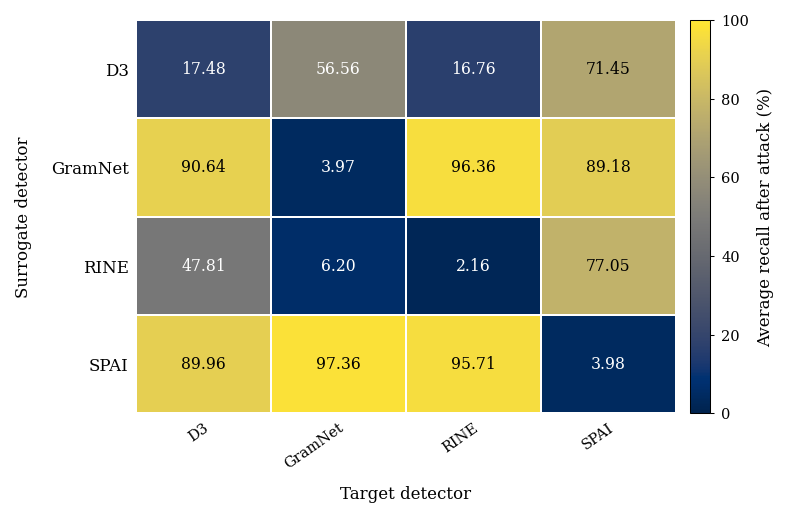}
    \caption{Tricker, unmatched generators.}
    \label{fig:transfer_detectors_tricker_unmatch}
  \end{subfigure}
  \begin{subfigure}[t]{.48\columnwidth}
    \centering
    \includegraphics[width=\linewidth]{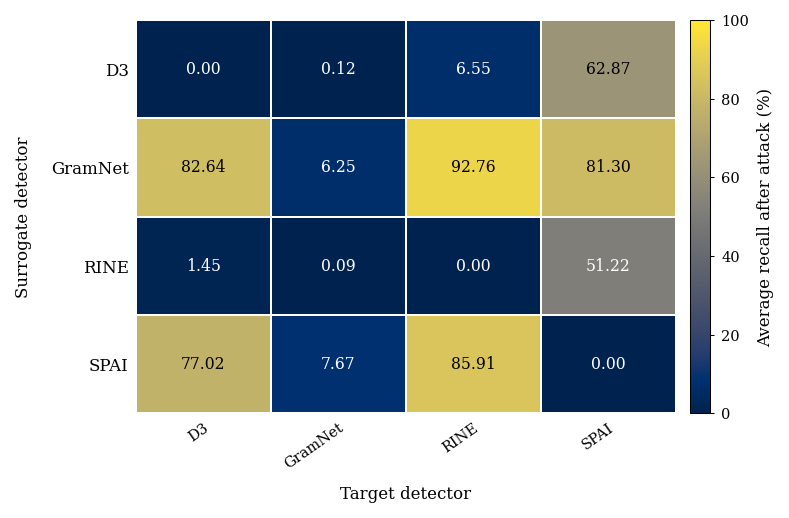}
    \caption{MIG}
    \label{fig:transfer_detectors_MIG}
  \end{subfigure}
  \begin{subfigure}[t]{.48\columnwidth}
    \centering
    \includegraphics[width=\linewidth]{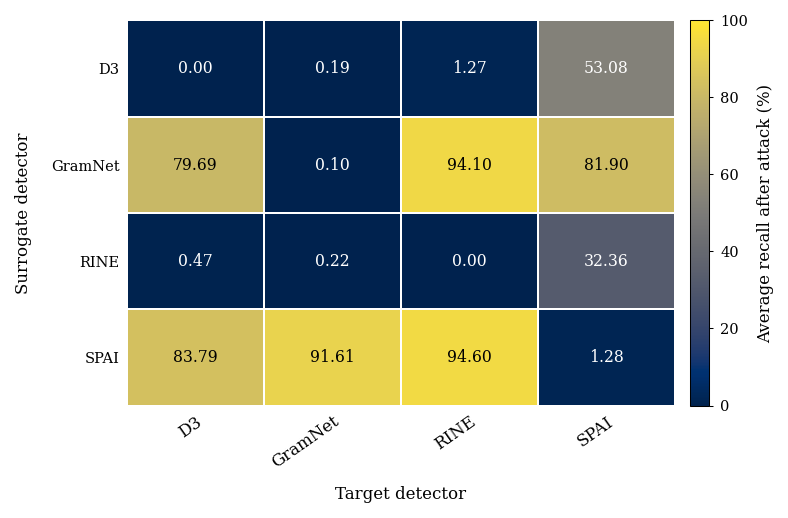}
    \caption{MuMoDIG}
    \label{fig:transfer_detectors_MuMoDIG}
  \end{subfigure}
  \caption{Cross-detector transferability. Rows are source detectors and columns are target detectors. Matrices reveal asymmetry.}
  \label{fig:across_detectors}
\end{figure}

Fig.~\ref{fig:across_detectors} compares cross-detector transferability across attacks. For Tricker, each cell corresponds to a source detector $D_s$, used to train the attack, and a target detector $D_t$, used only for evaluation. We consider two transfer settings. In the matched-generator setting, shown in Fig.~\ref{fig:transfer_detectors_tricker_match}, Tricker is trained on $(D_s, G_s)$ and evaluated on the target detector $D_t$ using images from the same generator $G_s$; each cell is averaged over the five surrogate generators. In the unmatched-generator setting, shown in Fig.~\ref{fig:transfer_detectors_tricker_unmatch}, Tricker is evaluated on unseen target generators $G_t \neq G_s$; each cell is therefore averaged over all source--target generator combinations, yielding $20$ values per detector pair. Lower recall indicates stronger transferability, and in the unmatched setting it reflects transfer across both detectors and generators.

For Tricker, the results reveal a strongly asymmetric transfer topology: successful transfer from $D_s$ to $D_t$ does not imply successful transfer in the reverse direction. In the unmatched-generator setting of Fig.~\ref{fig:transfer_detectors_tricker_unmatch}, $\mathrm{RINE} \rightarrow \mathrm{GramNet}$ yields $6.20\%$ recall, whereas $\mathrm{GramNet} \rightarrow \mathrm{RINE}$ yields $96.36\%$. This indicates that GramNet is a weak surrogate for attacking RINE. A similar asymmetry is observed between $\mathrm{D}^3$ and RINE in the full transferability setting, where the two directions achieve recalls of $16.76\%$ and $47.81\%$, respectively.

For perturbation-based methods, we evaluate detector transfer by crafting adversarial images against a source detector and testing them on a different target detector. These methods also exhibit asymmetric transferability, particularly between $\mathrm{D}^3$ and GramNet. However, unlike Tricker, they show stronger bidirectional transfer between $\mathrm{D}^3$ and RINE. Overall, MIG transfers better across detectors than MuMoDIG, but this improved transferability comes at the cost of lower adversarial image quality.



\begin{table*}[t]
\centering
\caption{
Post-processing in terms of mean fake-image recall
(\%). Cell backgrounds encode the recall change relative to the
original detector-averaged recall, $90.36\%$: deeper blue
denotes a larger recall reduction, gray denotes little change, and
orange denotes an increase. Lower recall indicates stronger
detector evasion. Colors represent total degradation relative to the original.
}
\label{tab:post_processing_main}

\begin{threeparttable}

\small
\setlength{\tabcolsep}{7.0pt}
\renewcommand{\arraystretch}{1.16}

\begin{tabular}{@{}llcccc@{}}
\toprule

\textbf{Detector setting}
&
\textbf{Attack}
&
\textbf{JPEG--75}
&
\shortstack{
  \textbf{Blur}\\
  $\boldsymbol{\sigma=2}$
}
&
\shortstack{
  \textbf{Downsample}\\
  $\boldsymbol{0.5\times}$
}
&
\shortstack{
  \textbf{Upsample}\\
  $\boldsymbol{2\times}$
}
\\

\midrule

PP only
&
No attack
&
\dcell{55.07}
&
\dcell{86.08}
&
\dcell{81.09}
&
\dcell{89.42}
\\

\midrule

\multirow{3}{*}{
  \shortstack[l]{
    Detector matched\\
    $D_s=D_t$
  }
}
&
Tricker
&
\dcell{32.15}
&
\dcell{59.24}
&
\dcell{24.67}
&
\dcell{13.91}
\\

&
MIG
&
\dcell{0.46}
&
\dcell{25.37}
&
\dcell{2.70}
&
\dcell{1.60}
\\

&
MuMoDIG
&
\dcell{2.78}
&
\dcell{37.09}
&
\dcell{1.46}
&
\dcell{0.44}
\\

\midrule

\multirow{3}{*}{
  \shortstack[l]{
    Detector transfer\\
    $D_s\neq D_t$
  }
}
&
Tricker
&
\dcell{54.78}
&
\dcell{80.98}
&
\dcell{71.11}
&
\dcell{72.47}
\\

&
MIG
&
\dcell{46.37}
&
\dcell{77.82}
&
\dcell{49.46}
&
\dcell{45.01}
\\

&
MuMoDIG
&
\dcell{45.80}
&
\dcell{80.24}
&
\dcell{55.86}
&
\dcell{53.90}
\\

\bottomrule
\end{tabular}

%


\end{threeparttable}
\end{table*}

\subsection{Post-processing operations}

Post-processing (PP) operations are common when images are uploaded to online platforms and can substantially affect detector performance. We therefore investigate how the studied attacks interact with four representative PP operations: JPEG compression with a quality factor of 75, Gaussian blurring with $\sigma=2$, downsampling to half the original resolution, and bicubic upsampling to twice the resolution.


Tab.~\ref{tab:post_processing_main} summarizes the performance of the evaluated attacks under different PP operations. The \textit{PP only} setting serves as a baseline, measuring the effect of each PP operation on detector performance without an attack and averages over the four detectors. In the \textit{detector-matched} setting, recall is measured on the same detector used to generate the attack and then averaged across the four source-detector configurations. In the \textit{transfer} setting, recall is averaged over the three detectors not involved in generating the attack and subsequently across all source detectors. Because Tricker is trained separately for each detector–generator pair, we additionally average its results across all source generators to obtain one value per source detector, enabling a direct comparison with MIG and MuMoDIG.


PP alone can degrade detector performance, although the magnitude of this effect varies across operations. JPEG causes the largest average recall reduction of \(35.29\%\), whereas upsampling has only a minor effect, reducing recall by \(0.94\%\). For all PP operations, combining PP with an attack yields lower recall on average than applying PP alone, demonstrating that Tricker, MIG, and MuMoDIG provide additional evasion beyond the degradation caused by PP. Nevertheless, PP generally weakens Tricker in the detector-matched setting, as all PP combinations produce higher recall than its white-box baseline of \(0.71\%\). In contrast, MIG and MuMoDIG largely retain their effectiveness under PP; notably, MIG combined with JPEG-75 achieves a recall of only \(0.46\%\), although Gaussian blurring noticeably reduces the effectiveness of both perturbation-based attacks.

In the transfer setting, JPEG compression improves Tricker's evasion, reducing its average recall from \(67.96\%\) in generator matched setting to \(54.78\%\). In contrast, the other PP operations increase recall and therefore weaken the attack. MIG remains comparatively stable under JPEG compression, downsampling, and upsampling, achieving recalls of \(46.37\%\), \(49.46\%\), and \(45.01\%\), respectively, all close to its original transfer recall of \(45.8\%\). Gaussian blurring is the main exception, substantially reducing MIG's effectiveness. MuMoDIG exhibits a similar pattern, although it benefits more noticeably from JPEG compression, which reduces its recall from \(51.1\%\) to \(45.8\%\). Overall, MIG and MuMoDIG consistently achieve stronger evasion than Tricker in both the detector-matched and transfer settings, as indicated by their darker table cells.


An important pattern is obscured by the aggregation in Table~\ref{tab:post_processing_main}: the underlying transferability topology persists after PP. Rather than affecting all source-target pairs uniformly, PP magnifies differences in their existing transferability relative to the PP-only baseline. For highly transferable pairs, it further reduces recall and strengthens evasion; for weakly transferable pairs, it increases recall and weakens evasion. This behavior is clearly visible in the disaggregated results reported in \autoref{app:post}, which provides the complete tables and additional analysis.

\section{Limitations and responsible use}
\label{limitations}
Our study is intended as a red-team evaluation of synthetic-image detectors rather than as a tool for misuse. The attack assumes gradient access to a differentiable surrogate detector during training and evaluates transfer-based black-box behavior rather than query-based adaptation to a deployed target. Our main experiments are conducted at $256 \times 256$ resolution, and we do not model watermarking or platform-specific transcoding. In addition, PSNR, SSIM, and RMSE capture low-level fidelity, whereas LPIPS and FID capture higher-level perceptual similarity; none fully replaces human perceptual studies. We also do not evaluate adaptive defenses, adversarial training, or detector ensembles. 

\section{Conclusion}
\label{conc}
We systematically evaluated the robustness of synthetic-image detectors under white-box and transferable attacks, comparing the perturbation-based MIG and MuMoDIG attacks with the amortized generative attack Tricker across four detectors and five generators. Tricker achieves strong white-box evasion with high perceptual quality, limited training data, and efficient single-pass inference. Our two-axis evaluation shows that transferability depends jointly on the surrogate--target detector pair and the generator distribution. MIG and MuMoDIG are particularly sensitive to specific detector--generator combinations, whereas Tricker generally provides more stable performance for image quality and white-box attack. We also find that common post-processing operations magnify the transferability topology. Finally, although Tricker requires one-time training, its amortized cost becomes advantageous at scale. Overall, reliable robustness evaluation should jointly consider detector and generator transfer, perceptual quality, post-processing effects, and attack-generation cost.


{
    \small
    \bibliographystyle{ieeenat_fullname}
    \bibliography{main}
}

\clearpage
\appendix
\section{Complementary Related Work}
\label{app:com_related_work}

\subsection{Generators} 
Advances in image generation have substantially improved visual fidelity and realism, but they also introduce distribution shifts that challenge existing AI-generated image detectors. In particular, state-of-the-art detectors can degrade sharply on images from newly developed generators, as each generation process may introduce distinct artifacts and frequency-domain fingerprints~\cite{corvi2023detection,bammey2023synthbuster,schinas2024sidbench}. To study how the generation process influences attack transferability, we select five representative generators spanning different model families and generation paradigms: BigGAN~\cite{brock2018large}, VQDM~\cite{gu2022vector}, SDV1.5~\cite{rombach2022high}, VAR~\cite{tian2024visual}, and SDV3.5~\cite{esser2024scaling}.

BigGAN is a large-scale class-conditional GAN for image synthesis that scales model capacity and batch size while using architectural changes, sampling techniques, and training-stability improvements, trained on ImageNet~\cite{deng2009imagenet}. 

VQDM is a vector-quantized diffusion model for text-to-image generation that models VQ-VAE latent tokens with a conditional DDPM~\cite{ho2020denoising}. It uses a mask-and-replace diffusion strategy to reduce accumulated prediction errors; released pretrained variants include MSCOCO~\cite{lin2014microsoft} and ImageNet. 

SDV1.5 is a latent text-to-image diffusion model that combines a downsampling-factor-8 autoencoder with a U-Net denoiser conditioned by a fixed CLIP ViT-L/14 text encoder. It is initialized from Stable Diffusion v1.2 and fine-tuned for 595k steps at 512×512 resolution on LAION-Aesthetics v2~\cite{schuhmann2022laionaestheticsv2}. 

VAR is a GPT-style~\cite{radford2018improving} visual autoregressive Transformer that generates images through coarse-to-fine next-scale prediction instead of standard raster-scan next-token prediction. It is trained on ImageNet~\cite{deng2009imagenet}. 

SDV3.5 is a Stable Diffusion 3.5 text-to-image diffusion model family based on the Multimodal flow matching Transformer architecture, using CLIP and T5 text encoders~\cite{radford2021learning,raffel2020exploring} with QK normalization. It is trained on a mixture of synthetic data and filtered publicly available data.

\subsection{Detectors} Detectors can be categorized along two main axes: the input representation they exploit, such as raw pixels or fingerprint-based features, and the backbone architecture they employ, such as convolutional or transformer-based models~\cite{schinas2024sidbench}. To study how adversarial vulnerabilities are shared across these categories, we select four representative detectors: GramNet~\cite{liu2020global}, RINE~\cite{koutlis2024leveraging}, $\mathrm{D}^3$~\cite{yang2025d}, and SPAI~\cite{karageorgiou2025any}.

GramNet is a ResNet18-based detector~\cite{he2016deep} that exploits global image texture representations derived from Gram matrices for robust fake image detection. It is trained on StyleGAN images~\cite{karras2020analyzing}. 

RINE is a projection-based neural detector with a trainable importance estimator that leverages image representations extracted from intermediate Transformer blocks of CLIP’s ViT-L/14 image encoder~\cite{radford2021learning}. It is trained on Latent Diffusion images. 

$\mathrm{D}^3$ is a CLIP ViT-L/14-based detector that extracts features from both the original image and a patch-shuffled version, then uses self-attention to model the discrepancy between these feature representations before applying a linear classifier for real/fake prediction. It is trained in a multi-generator setting. 

SPAI is a ViT-B/16-based spectral detector~\cite{dosovitskiy2020image} that models the spectral distribution of real images through masked frequency reconstruction. It detects synthetic images using spectral reconstruction similarity and spectral context attention, while supporting any-resolution inference. Its spectral backbone is pretrained on ImageNet, and the detector is trained on real and Latent Diffusion images.

\section{Implementation Details}
\label{app:impl}

\paragraph{Tricker architecture.}
Tricker is implemented as a residual-attention U-Net-style autoencoder operating on
$256 \times 256$ RGB images. The encoder uses a base channel width of $32$ and
channel multipliers $(1,2,4,8)$, giving feature widths $32$, $64$, $128$, and $256$.
Each stage contains two residual blocks. A residual block consists of Group Normalization,
SiLU activation, and $3 \times 3$ convolutional layers. Downsampling is performed using
stride-$2$ convolution. The decoder mirrors the encoder using bilinear upsampling followed
by residual refinement, and skip connections concatenate encoder features with decoder
features at the corresponding resolution. Self-attention blocks are inserted at the
$128$-channel stage, the $256$-channel stage, the bottleneck, and the first decoder stage.
The final prediction head applies residual refinement followed by Group Normalization,
SiLU activation, a $3 \times 3$ convolution, and a sigmoid activation so that the output
image lies in $[0,1]$. The resulting model has approximately $8$M parameters.

\paragraph{Training objective.}
Let $D_s$ denote the frozen surrogate detector and let
$\tilde{x}=T_{\theta}(x)$ be the Tricker output for an input synthetic image $x$.
All reported experiments use a hard real-class target
\[
q = [1,0],
\]
where the class order is $[\mathrm{real}, \mathrm{fake}]$. The training objective is
\[
\mathcal{L}(\theta)
=
2 \, D_{\mathrm{KL}}
\left(
q \,\middle\|\, p_{D_s}(\cdot \mid \tilde{x})
\right)
+
5 \, \mathrm{RMSE}_{m}(x,\tilde{x}),
\]
where $p_{D_s}(\cdot \mid \tilde{x})$ is the surrogate detector output distribution and
$m$ is the valid-image mask. The masked RMSE for one image is computed in normalized
$[0,1]$ space as
\[
\mathrm{RMSE}_{m}(x,\tilde{x})
=
\sqrt{
\frac{
\sum_{c,i,j} m_{i,j}
\left(\tilde{x}_{c,i,j}-x_{c,i,j}\right)^2
}{
3\sum_{i,j} m_{i,j}
}
}.
\]
The denominator of RMSE acts as a scaling factor for $\beta = 5$ so it matches our discussion of small $\beta$ to achieve high attack success rate. The detector temperature is set to $1.0$, and the detector loss is active from the first epoch with no ramp-up. 

\paragraph{Optimization.}
Tricker is trained with AdamW using learning rate $10^{-4}$, weight decay $10^{-4}$,
and betas $(0.9,0.95)$. The batch size is $8$. Training runs for at most $60$ epochs and
uses early stopping with patience $4$ based on the validation loss. Gradients are clipped
to norm $0.5$. The learning-rate schedule uses a linear warm-up for the first $3\%$ of
training steps followed by cosine decay to $5\%$ of the initial learning rate. An exponential
moving average of the model weights is maintained with decay $0.999$ and is used for
evaluation. The random seed is fixed to $42$.

\paragraph{Data.}
For each generator, we use $6{,}000$ synthetic images: $4{,}000$ for training and 2,000 images for attack evaluation. No held-out image is used for gradient updates of Tricker. In the white-box setting, these results should be interpreted as held-out validation attack performance under a standard adaptive white-box threat model. In generator-transfer experiments, the target generator is excluded from Tricker training, so evaluation is performed on a generator distribution unseen during attack training. All images are processed at $256 \times 256$ resolution. The training loader applies horizontal flipping with probability $0.5$. The generators used in the experiments are BigGAN, VQDM, SDV1.5, VAR, and SDV3.5.

\paragraph{Surrogate detector preprocessing.}
Each surrogate detector is kept frozen during Tricker training. Detector-specific
preprocessing is applied inside the differentiable detector wrapper before forwarding
$\tilde{x}$ through the detector. For GramNet, images are resized to $256 \times 256$,
center-cropped to $224 \times 224$, and normalized using ImageNet statistics:
\[
\mu_{\mathrm{IN}}=(0.485,0.456,0.406), \qquad
\sigma_{\mathrm{IN}}=(0.229,0.224,0.225).
\]
For RINE and D$^3$, the same spatial preprocessing is used, but images are normalized
using CLIP statistics:
\[
\mu_{\mathrm{CLIP}}=(0.48145466,0.4578275,0.40821073),
\]
\[
\sigma_{\mathrm{CLIP}}=(0.26862954,0.26130258,0.27577711).
\]
For SPAI, we use the original SPAI configuration file. This configuration performs normalization to $[0-1]$ range, activates SPAI's arbitrary resolution, and performs padding if a dimension is smaller than 224. 
The image is then passed to SPAI without ImageNet or CLIP normalization. Since SPAI uses
arbitrary-resolution inference, the preprocessed images are provided as a list
of tensors, each of shape $[1,3,H,W]$, and the SPAI model processes this list
using the configured feature-extraction batch size of $400$.

For detectors that return a single
binary logit $z$, we convert it to two-class logits $[0,z]$, where the second class is the
fake class.

\paragraph{Validation metrics.}
Detector evasion is evaluated at threshold $0.5$ on the real-class probability. An input is
counted as fake before attack when
\[
p_{D_s}(\mathrm{real}\mid x) < 0.5,
\]
and it is counted as successfully attacked when
\[
p_{D_s}(\mathrm{real}\mid T_{\theta}(x)) \geq 0.5.
\]


Because our validation data are sampled randomly, each detector exhibits a different baseline recall and therefore identifies a different number of fake images before the attack. To provide a fair and transparent evaluation, we report both the pre-attack and post-attack recall for each detector. We use the recall drop as the primary indicator of attack success, since it directly measures how much the attack reduces the detector's ability to identify synthetic images relative to its own baseline performance.


\paragraph{Image-quality metrics.}
Image quality is measured between the original synthetic image $x$ and the Tricker output
$\tilde{x}$. We report PSNR, SSIM, and RMSE as low level perceptual metrics while FID and LPIPS for high level. RMSE is computed in $[0,255]$
space.

\section{SIDBench Detectors Recall}
\label{app:base}

\begin{figure}
    \centering
    \includegraphics[width=\linewidth]{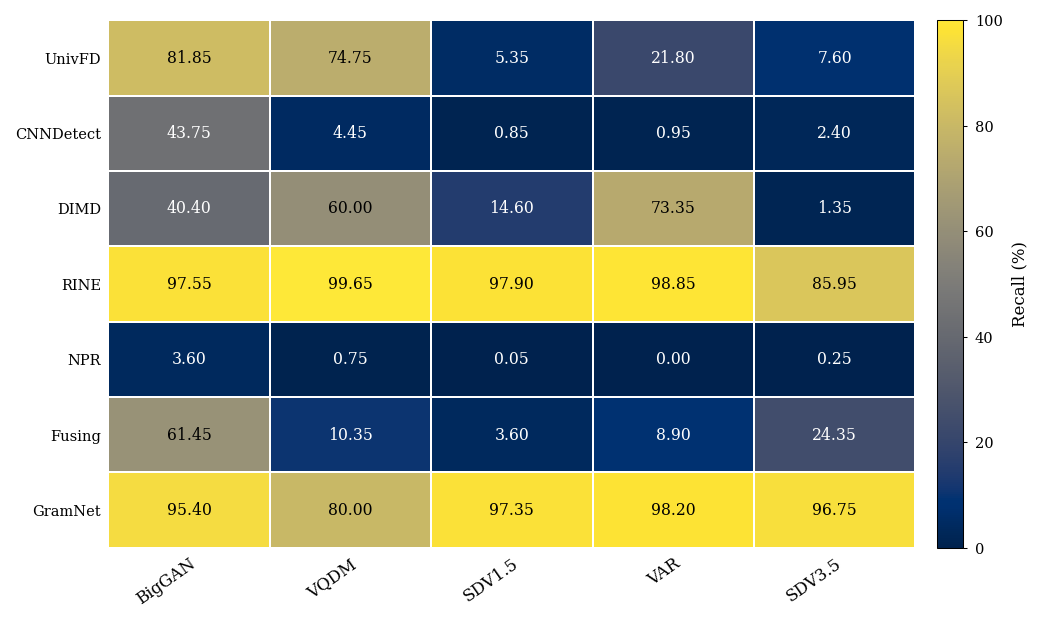}
    \caption{SIDBench detectors recall without attack.}
    \label{fig:all_detectors}
\end{figure}

Fig.~\ref{fig:all_detectors} reports the baseline recall of the evaluated SIDBench detectors on our validation data. Several of these detectors have low recall on our generator set, making them less informative for adversarial evaluation. In particular, when a detector already fails to reliably identify synthetic images before the attack, any subsequent recall drop provides a weak and potentially misleading signal of attack success. We therefore retain RINE and GramNet from the original SIDBench detector set and additionally integrate $\mathrm{D}^3$ and SPAI as recent detectors with complementary architectures and stronger relevance to current SID robustness evaluation.

\section{Positioning with PGD}
\label{app:pgd}

\begin{figure}[h]
    \centering
    \includegraphics[width=\linewidth]{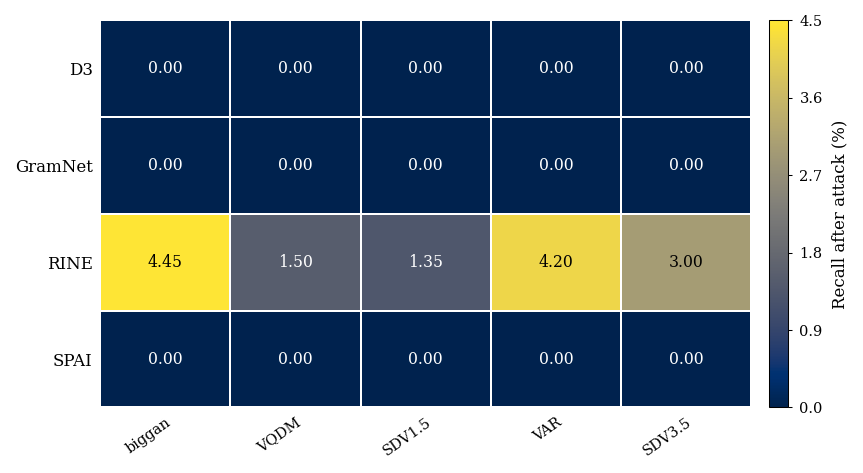}
    \caption{white-box attack PGD}
    \label{fig:pgd}
\end{figure}

Using white-box PGD with an $\ell_\infty$ perturbation budget of $\varepsilon = 16/255$, the average recall is reduced to $2.9\%$. PGD drives recall to $0\%$ for all detectors across all generators, with RINE as the only exception as you can see in Fig. ~\ref{fig:pgd}. For RINE, recall varies substantially across generators, mirroring the behavior observed for MIG and MuMoDIG. These results reinforce our finding that perturbation-based attacks are more sensitive to the choice of detector--generator pair, whereas Tricker achieves consistently strong evasion across all evaluated pairs.


\section{Artifacts discussion}
\label{app:artifacts}

\begin{figure*}[t]
    \centering
    \begingroup
    \setlength{\tabcolsep}{1.2pt}
    \renewcommand{\arraystretch}{0.95}
    \scriptsize
    \newcommand{\qualimgw}{0.176\linewidth}
    \newcommand{\qimg}[1]{\includegraphics[width=\qualimgw]{figs/#1}}
    \newcommand{\rowlab}[1]{%
        \raisebox{\dimexpr\qualimgw/2-\height/2\relax}{\rotatebox{90}{\textbf{#1}}}%
    }

    \begin{tabular}{@{}r@{\hspace{3pt}}ccccc@{}}
        & \textbf{BigGAN} & \textbf{SDV1.5} & \textbf{SDV3.5} & \textbf{VAR} & \textbf{VQDM} \\

        \rowlab{Original} &
        \qimg{original_bigGAN_worst.png} &
        \qimg{original_SDV1.5_worst.png} &
        \qimg{original_SDV3.5_worst.png} &
        \qimg{original_VAR_worst.png} &
        \qimg{original_VQDM_worst.png} \\

        \rowlab{Tricker} &
        \qimg{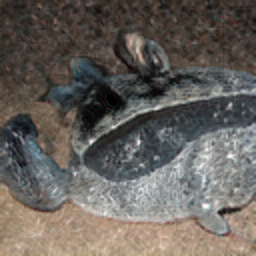} &
        \qimg{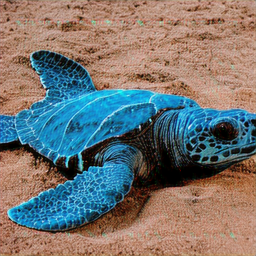} &
        \qimg{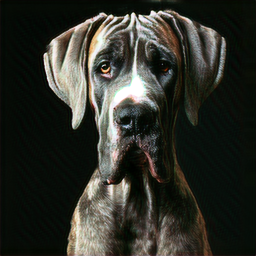} &
        \qimg{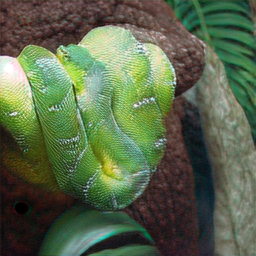} &
        \qimg{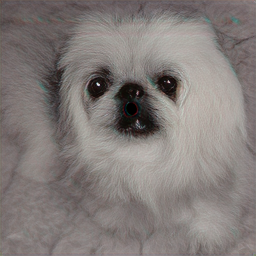} \\

        \rowlab{MIG} &
        \qimg{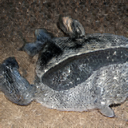} &
        \qimg{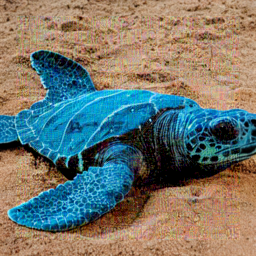} &
        \qimg{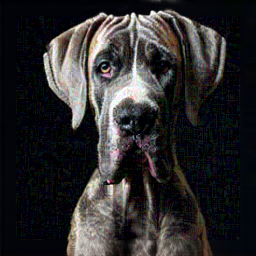} &
        \qimg{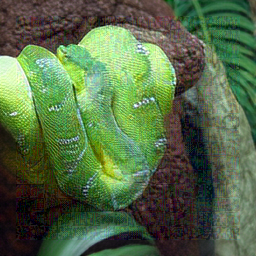} &
        \qimg{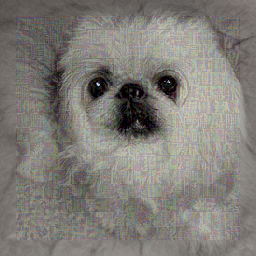} \\

        \rowlab{MuMoDIG} &
        \qimg{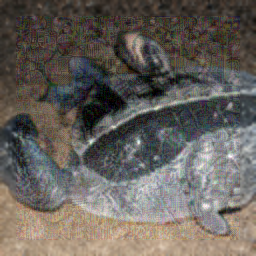} &
        \qimg{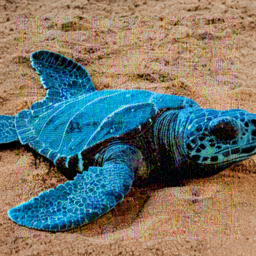} &
        \qimg{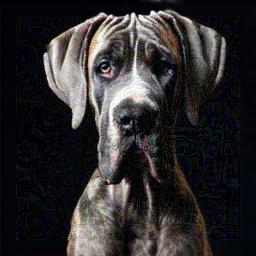} &
        \qimg{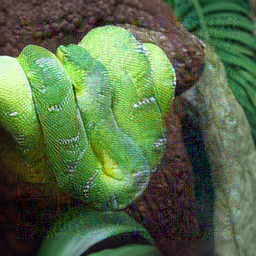} &
        \qimg{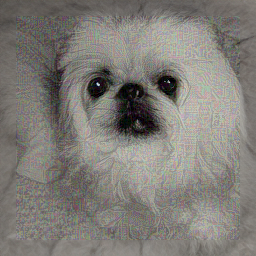}
    \end{tabular}
    \endgroup

    \vspace{10pt}
    \caption{Qualitative worst-case examples generated with D3.}
    \label{fig:qualitative-worst-cases}
\end{figure*}

Figure~\ref{fig:qualitative-worst-cases} shows qualitative worst-case examples generated with $\mathrm{D}^3$ across the five generator families, comparing Tricker with MIG and MuMoDIG. Tricker can introduce visible but localized artifacts, such as ring-like patterns, while generally preserving object identity and scene layout. For example, the SDV1.5 sea turtle remains recognizable, although the sand background becomes slightly reddish and noisy. The SDV3.5 dog preserves its pose and facial structure, while some local regions become over-brightened. The VAR snake remains comparatively stable, with only mild localized artifacts, and the VQDM dog shows a small dark artifact below the nose.

Compared with Tricker, MIG and MuMoDIG often introduce stronger global high-frequency degradation in these examples. MIG produces more noticeable texture corruption in the SDV3.5 and VQDM dog images, while MuMoDIG introduces amplified noise in the BigGAN turtle, stronger texture distortion in the SDV1.5 sand, and grid-like artifacts in the VQDM fur. Overall, these examples suggest that Tricker tends to preserve semantic content more consistently, whereas perturbation-based attacks can produce more spatially distributed noise.

\section{ All details for Post-processing survivability}
\label{app:post}

Images distributed through online platforms are frequently recompressed, blurred, or resized, and these operations may affect both the adversarial modification and the detector response. We therefore apply JPEG recompression at quality factor $Q=75$, Gaussian blur with $\sigma=2$, $0.5\times$ downsampling, and $2\times$ bicubic upsampling after attack generation, and report the resulting fake-image recall in Table~\ref{tab:post_processing_all_attacks}. Each Tricker entry averages five source-generator-specific attack models, each evaluated on five target-generator distributions, resulting in 25 source--target generator combinations per cell. The no-attack, MIG, and MuMoDIG entries are averaged over the five target-generator distributions. Importantly, the no-attack controls demonstrate that post-processing is not neutral: JPEG--75 reduces GramNet's recall from $R_0=93.54\%$ to $0.48\%$, whereas Gaussian blur increases the recalls of GramNet and SPAI to $98.67\%$ and $92.66\%$, respectively. Consequently, the attacked results should be interpreted relative to the corresponding post-processed no-attack condition, in addition to the original unprocessed recall $R_0$. In the matched-detector setting ($D_s=D_t$, shown in bold), MIG and MuMoDIG yield low mean recalls across the four target detectors after JPEG recompression ($0.46\%$ and $2.78\%$), downsampling ($2.70\%$ and $1.46\%$), and upsampling ($1.60\%$ and $0.44\%$), respectively. For Tricker, the mean matched recalls are $32.16\%$, $59.24\%$, $24.67\%$, and $13.91\%$ under JPEG recompression, blurring, downsampling, and upsampling, respectively; nevertheless, Tricker reduces recall relative to the corresponding post-processed no-attack condition for every matched detector. Finally, transfer effectiveness remains strongly dependent on the source--target detector pair and is generally weaker than in the matched setting. In particular, the near-zero GramNet recalls obtained after JPEG recompression should not be attributed solely to the attacks, since JPEG--75 alone reduces GramNet's recall to nearly zero.

\begin{table*}[p]
\centering

\caption{
Fake-image recall after applying common post-processing. Results are grouped by attack family and surrogate
detector. Cell colors are computed independently for each target
detector relative to its original unprocessed recall $R_0$: values
near $R_0$ are gray, lower values become increasingly blue, and
higher values become increasingly red. Lower recall indicates
stronger detector evasion.
}
\label{tab:post_processing_all_attacks}

\begin{threeparttable}

\footnotesize
\setlength{\tabcolsep}{1pt}
\renewcommand{\arraystretch}{0.86}

\begin{tabular}{@{}llcccc@{}}
\toprule

\shortstack[l]{
    \textbf{Method applied}\\
    \textbf{before post-processing}
}
&
\shortstack{
    \textbf{Post-}\\
    \textbf{processing}
}
&
\multicolumn{4}{c}{\textbf{Target detector} $D_t$}
\\

\cmidrule(lr){3-6}

&
&
\shortstack{
    \textbf{D$^{3}$}\\[-1pt]
    {\tiny $R_0=87.57\%$}
}
&
\shortstack{
    \textbf{GramNet}\\[-1pt]
    {\tiny $R_0=93.54\%$}
}
&
\shortstack{
    \textbf{RINE}\\[-1pt]
    {\tiny $R_0=95.98\%$}
}
&
\shortstack{
    \textbf{SPAI}\\[-1pt]
    {\tiny $R_0=84.34\%$}
}
\\

\midrule


\multirow{4}{*}{\textbf{No attack}}
&
\PlainPPRow
    {JPEG--75}
    {63.18}
    {0.48}
    {72.87}
    {83.75}
\\

&
\PlainPPRow
    {Blur ($\sigma=2$)}
    {74.12}
    {98.67}
    {78.87}
    {92.66}
\\

&
\PlainPPRow
    {Downsample ($0.5\times$)}
    {77.16}
    {83.01}
    {81.84}
    {82.34}
\\

&
\PlainPPRow
    {Upsample ($2\times$, bicubic)}
    {85.31}
    {93.26}
    {93.95}
    {85.17}
\\

\familyrule


\multirow{4}{*}{
    \shortstack[l]{
        \textbf{Tricker}\\
        ($D_s=\mathrm{D}^{3}$)
    }
}
&
\DthreeSourceRow
    {JPEG--75}
    {40.22}
    {0.25}
    {53.80}
    {84.66}
\\

&
\DthreeSourceRow
    {Blur ($\sigma=2$)}
    {61.24}
    {95.92}
    {61.23}
    {90.05}
\\

&
\DthreeSourceRow
    {Downsample ($0.5\times$)}
    {40.43}
    {55.51}
    {34.06}
    {78.89}
\\

&
\DthreeSourceRow
    {Upsample ($2\times$, bicubic)}
    {28.30}
    {61.34}
    {25.69}
    {76.88}
\\

\sourcerule


\multirow{4}{*}{
    \shortstack[l]{
        \textbf{Tricker}\\
        ($D_s=\mathrm{GramNet}$)
    }
}
&
\GramSourceRow
    {JPEG--75}
    {62.98}
    {0.11}
    {79.61}
    {90.47}
\\

&
\GramSourceRow
    {Blur ($\sigma=2$)}
    {73.06}
    {79.39}
    {83.68}
    {95.56}
\\

&
\GramSourceRow
    {Downsample ($0.5\times$)}
    {79.93}
    {18.01}
    {86.39}
    {95.33}
\\

&
\GramSourceRow
    {Upsample ($2\times$, bicubic)}
    {84.05}
    {9.79}
    {92.68}
    {94.14}
\\

\sourcerule


\multirow{4}{*}{
    \shortstack[l]{
        \textbf{Tricker}\\
        ($D_s=\mathrm{RINE}$)
    }
}
&
\RineSourceRow
    {JPEG--75}
    {56.92}
    {0.14}
    {50.40}
    {84.41}
\\

&
\RineSourceRow
    {Blur ($\sigma=2$)}
    {65.28}
    {70.44}
    {54.33}
    {94.13}
\\

&
\RineSourceRow
    {Downsample ($0.5\times$)}
    {60.70}
    {19.73}
    {26.36}
    {89.06}
\\

&
\RineSourceRow
    {Upsample ($2\times$, bicubic)}
    {58.64}
    {12.29}
    {8.40}
    {86.76}
\\

\sourcerule


\multirow{4}{*}{
    \shortstack[l]{
        \textbf{Tricker}\\
        ($D_s=\mathrm{SPAI}$)
    }
}
&
\SPAISourceRow
    {JPEG--75}
    {66.94}
    {1.58}
    {75.61}
    {37.89}
\\

&
\SPAISourceRow
    {Blur ($\sigma=2$)}
    {71.01}
    {99.78}
    {71.64}
    {42.00}
\\

&
\SPAISourceRow
    {Downsample ($0.5\times$)}
    {83.70}
    {88.13}
    {81.89}
    {13.87}
\\

&
\SPAISourceRow
    {Upsample ($2\times$, bicubic)}
    {88.06}
    {97.89}
    {91.21}
    {9.14}
\\

\familyrule


\multirow{4}{*}{
    \shortstack[l]{
        \textbf{MIG}\\
        ($D_s=\mathrm{D}^{3}$)
    }
}
&
\DthreeSourceRow
    {JPEG--75}
    {0.80}
    {0.21}
    {35.23}
    {72.90}
\\

&
\DthreeSourceRow
    {Blur ($\sigma=2$)}
    {30.88}
    {98.63}
    {44.27}
    {99.59}
\\

&
\DthreeSourceRow
    {Downsample ($0.5\times$)}
    {3.81}
    {1.49}
    {19.14}
    {80.36}
\\

&
\DthreeSourceRow
    {Upsample ($2\times$, bicubic)}
    {0.00}
    {0.18}
    {7.23}
    {79.69}
\\

\sourcerule


\multirow{4}{*}{
    \shortstack[l]{
        \textbf{MIG}\\
        ($D_s=\mathrm{GramNet}$)
    }
}
&
\GramSourceRow
    {JPEG--75}
    {67.56}
    {0.01}
    {84.84}
    {86.08}
\\

&
\GramSourceRow
    {Blur ($\sigma=2$)}
    {70.68}
    {40.47}
    {80.53}
    {99.16}
\\

&
\GramSourceRow
    {Downsample ($0.5\times$)}
    {70.92}
    {6.40}
    {78.29}
    {87.71}
\\

&
\GramSourceRow
    {Upsample ($2\times$, bicubic)}
    {76.64}
    {6.39}
    {79.31}
    {85.47}
\\

\sourcerule


\multirow{4}{*}{
    \shortstack[l]{
        \textbf{MIG}\\
        ($D_s=\mathrm{RINE}$)
    }
}
&
\RineSourceRow
    {JPEG--75}
    {10.87}
    {0.21}
    {0.27}
    {62.06}
\\

&
\RineSourceRow
    {Blur ($\sigma=2$)}
    {27.57}
    {98.49}
    {4.81}
    {98.24}
\\

&
\RineSourceRow
    {Downsample ($0.5\times$)}
    {12.25}
    {1.10}
    {0.01}
    {69.14}
\\

&
\RineSourceRow
    {Upsample ($2\times$, bicubic)}
    {1.43}
    {0.13}
    {0.00}
    {64.73}
\\

\sourcerule


\multirow{4}{*}{
    \shortstack[l]{
        \textbf{MIG}\\
        ($D_s=\mathrm{SPAI}$)
    }
}
&
\SPAISourceRow
    {JPEG--75}
    {55.34}
    {0.11}
    {81.07}
    {0.77}
\\

&
\SPAISourceRow
    {Blur ($\sigma=2$)}
    {60.06}
    {99.51}
    {57.13}
    {25.33}
\\

&
\SPAISourceRow
    {Downsample ($0.5\times$)}
    {71.98}
    {40.35}
    {60.75}
    {0.56}
\\

&
\SPAISourceRow
    {Upsample ($2\times$, bicubic)}
    {75.26}
    {11.37}
    {58.68}
    {0.00}
\\

\familyrule


\multirow{4}{*}{
    \shortstack[l]{
        \textbf{MuMoDIG}\\
        ($D_s=\mathrm{D}^{3}$)
    }
}
&
\DthreeSourceRow
    {JPEG--75}
    {0.27}
    {0.28}
    {24.11}
    {73.57}
\\

&
\DthreeSourceRow
    {Blur ($\sigma=2$)}
    {32.35}
    {99.23}
    {43.89}
    {98.81}
\\

&
\DthreeSourceRow
    {Downsample ($0.5\times$)}
    {2.00}
    {3.77}
    {15.78}
    {74.35}
\\

&
\DthreeSourceRow
    {Upsample ($2\times$, bicubic)}
    {0.01}
    {0.35}
    {2.18}
    {77.44}
\\

\sourcerule


\multirow{4}{*}{
    \shortstack[l]{
        \textbf{MuMoDIG}\\
        ($D_s=\mathrm{GramNet}$)
    }
}
&
\GramSourceRow
    {JPEG--75}
    {68.74}
    {0.03}
    {86.04}
    {86.35}
\\

&
\GramSourceRow
    {Blur ($\sigma=2$)}
    {71.04}
    {58.12}
    {80.94}
    {99.09}
\\

&
\GramSourceRow
    {Downsample ($0.5\times$)}
    {72.41}
    {0.15}
    {81.35}
    {88.61}
\\

&
\GramSourceRow
    {Upsample ($2\times$, bicubic)}
    {77.83}
    {0.11}
    {84.11}
    {86.43}
\\

\sourcerule


\multirow{4}{*}{
    \shortstack[l]{
        \textbf{MuMoDIG}\\
        ($D_s=\mathrm{RINE}$)
    }
}
&
\RineSourceRow
    {JPEG--75}
    {6.24}
    {0.23}
    {0.01}
    {57.81}
\\

&
\RineSourceRow
    {Blur ($\sigma=2$)}
    {38.48}
    {99.38}
    {12.39}
    {98.10}
\\

&
\RineSourceRow
    {Downsample ($0.5\times$)}
    {12.39}
    {3.37}
    {0.02}
    {64.51}
\\

&
\RineSourceRow
    {Upsample ($2\times$, bicubic)}
    {0.40}
    {0.31}
    {0.00}
    {59.29}
\\

\sourcerule


\multirow{4}{*}{
    \shortstack[l]{
        \textbf{MuMoDIG}\\
        ($D_s=\mathrm{SPAI}$)
    }
}
&
\SPAISourceRow
    {JPEG--75}
    {64.00}
    {0.09}
    {82.15}
    {10.81}
\\

&
\SPAISourceRow
    {Blur ($\sigma=2$)}
    {65.97}
    {98.39}
    {69.61}
    {45.50}
\\

&
\SPAISourceRow
    {Downsample ($0.5\times$)}
    {79.40}
    {97.19}
    {77.18}
    {3.65}
\\

&
\SPAISourceRow
    {Upsample ($2\times$, bicubic)}
    {82.82}
    {93.91}
    {81.74}
    {1.62}
\\

\bottomrule
\end{tabular}






\end{threeparttable}
\end{table*}

\end{document}